\documentclass[conference]{IEEEtran}
\usepackage{cite}
\usepackage{amsmath,amssymb,amsfonts}
\usepackage{algorithmic}
\usepackage{graphicx}
\usepackage[round]{natbib}              
\usepackage{url}            
\graphicspath{{figures/}}
\usepackage{subcaption}
\usepackage{textcomp}
\usepackage{xcolor}
\usepackage{soul}
\usepackage{float}
\usepackage{geometry}
\usepackage{multirow}       
\usepackage{mathtools}      
\usepackage{fancyhdr}       
\usepackage{orcidlink}      
\usepackage{wrapfig}        
\usepackage{hyperref}       
\hypersetup{
  colorlinks= true, 
  urlcolor= magenta, 
  linkcolor= red, 
  citecolor= blue 
}

\def\BibTeX{{\rm B\kern-.05em{\sc i\kern-.025em b}\kern-.08em
    T\kern-.1667em\lower.7ex\hbox{E}\kern-.125emX}}

\makeatletter

\def\modelname{RGB-D-Fusion}

\title{\modelname: Image Conditioned Depth Diffusion of Humanoid Subjects}

\author{Sascha Kirch \orcidlink{0000-0002-5578-7555}$^{1}$, Valeria Olyunina$^{2}$, Jan Ondřej \orcidlink{0000-0002-5409-1521}$^{2}$, Rafael Pagés \orcidlink{0000-0002-5691-9580}$^{2}$, Sergio Martin \orcidlink{0000-0002-4118-0234}$^{1}$, Clara Pérez-Molina \orcidlink{0000-0001-8260-4155}$^{1}$ \\
	\normalsize $^{1}$UNED - Universidad Nacional de Educación a Distancia, Madrid, Spain\\
	\normalsize \href{mailto:skirch1@alumno.uned.es}{skirch1@alumno.uned.es}, 
	\{\href{mailto:smartin@ieec.uned.es}{smartin}, \href{mailto:clarapm@ieec.uned.es}{clarapm}\}@ieec.uned.es
	\\
	\normalsize $^{2}$Volograms ltd, Dublin, Ireland\\
	\normalsize \{\href{mailto:valeria@volograms.com}{valeria}, \href{mailto:jan@volograms.com}{jan}, \href{mailto:rafa@volograms.com}{rafa}\}@volograms.com
}

\begin{document}

\maketitle

\begin{abstract}
We present \modelname, a multi-modal conditional denoising diffusion probabilistic model to generate high resolution depth maps from low-resolution monocular RGB images of humanoid subjects. \modelname{} first generates a low-resolution depth map using an image conditioned denoising diffusion probabilistic model and then upsamples the depth map using a second denoising diffusion probabilistic model conditioned on a low-resolution RGB-D image. We further introduce a novel augmentation technique, depth noise augmentation, to increase the robustness of our super-resolution model.

\end{abstract}
\section{Introduction}\label{sec:introduction}

From immersive communications, games, virtual production, to virtual try-on and virtual fitting rooms, having an accurate 3D presentation of the body of a person is fundamental. For this, one would typically need to use a professional 3D capture stage with many cameras pointing at the person in the center of the stage, and then put all the videos together using a 3D reconstruction algorithm, such as the ones proposed by \cite{guo2019relightables}, \cite{collet2015high} or \cite{pages2018affordable}. These setups are complex (huge amount of data to manage, complicated calibration processes, high processing times) and expensive (a high number of cameras, professional lighting, computers, GPUs, networks and more), so to overcome these challenges, modern deep learning techniques aim to simplify the capture process by replacing the multi-camera setup with a single camera, from a single viewpoint. For example, \cite{saito2019pifu}, \cite{saito2020pifuhd}, and \cite{xiu2022icon} generate a full body 3D model of a person, \cite{escribano2022texture} propose a way to generate the textures of occluded areas. However, all these techniques output relatively low-resolution considering the target application, and they struggle with complicated poses where depth is difficult to convey from a single viewpoint.

To overcome these challenges, some approaches include additional information into the capturing process, including depth, which typically comes from a consumer-level depth sensor. Although new generation consumer-level depth sensors have significantly improved during the last few years, they still present high levels of noise and sometimes fail to capture details when the subjects are at a distance where their body can be fully captured (at around 2m and farther), and they are still sensitive to sun light (which interferes with the infra-red light these sensors typically use), dark materials (which absorb infra-red light) and non-Lambertian surfaces. 

In this work, we propose a new approach to generate high-resolution depth maps of humans using a single low-resolution image as input, using denoising diffusion probabilistic models (DDPMs). Our multi-modal DDPM was trained using high-resolution depth maps, extracted from a large dataset of photorealistic 3D models captured by Volograms \citep{volograms2021}, which allows us to obtain depth maps with higher accuracy than consumer-level depth sensors.

\begin{figure}[t]
  \centering
  \includegraphics[width=0.99\linewidth]{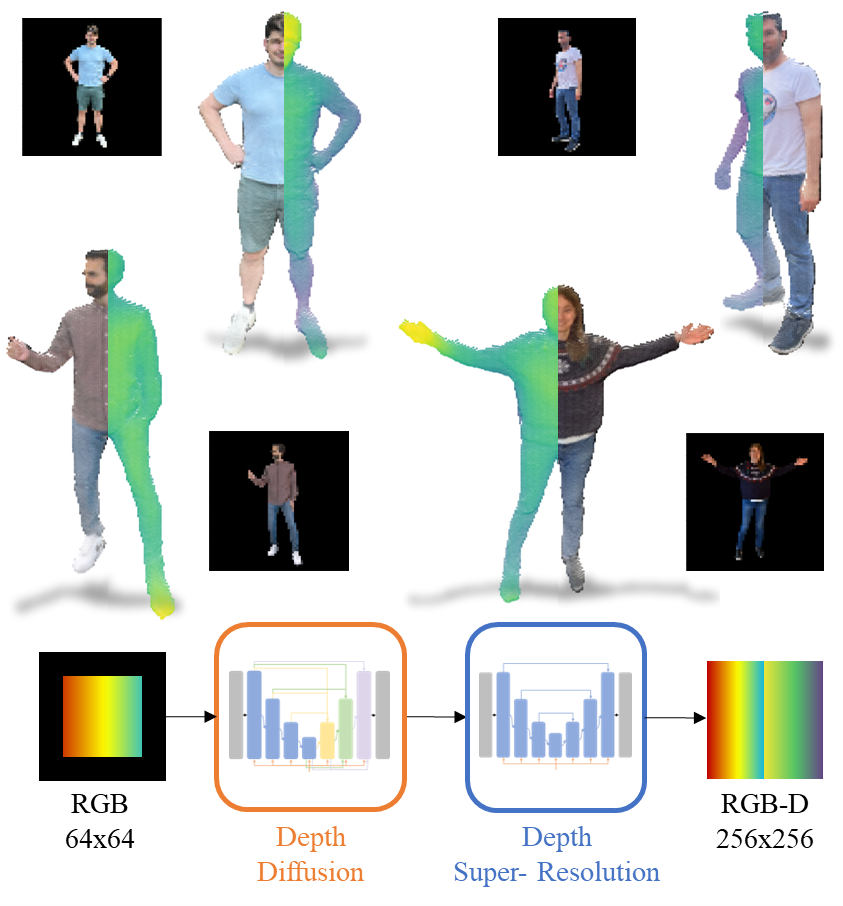}
  \caption{Our \modelname{} framework generates a high-resolution RGB-D image from a low-resolution RGB image of humanoid subjects. First, a low resolution depth map is generated using a conditional DDPM. Then, the depth is upsampled to a higher resolution using a second DDPM.}
  \label{fig:featured_image}
\end{figure}

Following the taxonomy for multi-modal deep learning systems by \cite{zhan_multimodal_2022} we frame \modelname{} as a multi-modal generative translation task using a joint data representation of our input modalities RGB and depth. No alignment is performed between these modalities since our data is perfectly aligned during generation. We further perform parallel data co-learning since our RGB and depth pairs inherently share a direct correspondence between their instances.

We summarize our main contributions of this paper as follows:
\begin{enumerate}
  \item We provide a framework for high resolution dense monocular depth estimation using diffusion models.
  \item We perform super-resolution for dense depth data conditioned on a multi-modal RGB-D input condition using diffusion models. 
  \item We introduce a novel augmentation technique, namely depth noise, to enhance the robustness of the depth super-resolution model.
  \item We perform rigorous ablations and experiments to validate our design choices.
\end{enumerate}

\section{Related Work} \label{sec:related_work}
In this section, we situate our work within a broader context and offer a concise summary of other relevant studies.

\subsection{Denoising Diffusion Probabilistic Models} Denoising diffusion models are a type of generative deep learning models first formulated by \cite{sohl-dickstein_deep_2015} and further extended to Denoising Diffusion Probabilistic models (DDPMs) by \cite{ho_denoising_2020,nichol_improved_2021}. These models use a Markov Chain to gradually convert one simple and well-known distribution (e.g. a Gaussian distribution) into a usually more complex data distribution the model can sample from. 

Diffusion models have been applied to many applications and modalities including image generation \citep{dhariwal_diffusion_2021}, image-to-image translation \citep{saharia_palette_2022}, super-resolution \citep{ho_cascaded_2021, saharia_image_2021}, video \citep{ho_video_2022,harvey_flexible_2022,yang_diffusion_2022}, audio \citep{kong_diffwave_2021,lee_priorgrad_2022}, text-to-image \citep{ramesh_hierarchical_2022, nichol_towards_2022,saharia_photorealistic_2022}, and simultaneous multi-modal generation \citep{ruan_mm-diffusion_2023}.

In the generative learning trilemma formulated by \cite{xiao_tackling_2022}, which states that generative models cannot satisfy all three key requirements fast sampling, high quality samples and mode coverage, diffusion models have shown good results in high quality image generation \citep{rampas_novel_2023,ho_cascaded_2021} and mode coverage. While Variational Auto-Encoders (VAEs) \citep{kingma_auto-encoding_2022,razavi_generating_2019} and flow based models \citep{dinh_density_2017,durkan_neural_2019} are strong in covering multi-modal data distributions and can be sampled from very fast, the quality of their samples is usually not as high as of Generative Adversarial Models (GANs) or diffusion models. GANs \citep{goodfellow_generative_2014,brock_large_2019,kirch_vologan_2022} on the other-hand produce high quality images and are sampled quickly but are prone to mode collapse and often do not cover the entire data distribution.

DDPMs require many reverse diffusion steps (often several hundreds or even thousands of steps) to sample from, making it difficult train and deploy them even on modern GPUs. Hence it is no surprise that a lot of research focuses on increasing the sample speed of diffusion models by reducing the number of required steps \citep{song_denoising_2022, xiao_tackling_2022, nichol_improved_2021, salimans_progressive_2022,chung_come-closer-diffuse-faster_2022}, perform diffusion in the latent space rather than the full-resolution data distribution \citep{preechakul_diffusion_2022,rombach_high-resolution_2022,sinha_d2c_2021, gu_vector_2022, tang_improved_2023} or formulate the diffusion problem as time-continuous problem to take advantage of accelerated stochastic differential equations (SDE) solvers \citep{song_score-based_2021,song_maximum_2021, song_improved_2020,karras_elucidating_2022}.

To control the output of the model it must be provided with an additional condition input. The model might be conditioned on another input of the same modality like in colorization \citep{saharia_palette_2022}, inpaintings \citep{batzolis_conditional_2021}, generation from semantic mask \citep{meng_sdedit_2022} or image super-resolution \citep{saharia_image_2021,ho_cascaded_2021}, on an input of different modality like in depth estimation \citep{duan_diffusiondepth_2023} or segmentation \citep{baranchuk_label-efficient_2022}, on class labels \citep{dhariwal_diffusion_2021} or on text prompts \citep{ramesh_hierarchical_2022, nichol_towards_2022,saharia_photorealistic_2022}; to name a few.

Depending on the representation of the condition input, it might be concatenated with the noise input \citep{batzolis_conditional_2021}, fed to multiple layers within the network like adaptive instance normalization \citep{karras_style-based_2019} or via an independent network branch \citep{zhang_adding_2023}. Beside the architectural choice one must also decide how strong the condition should be. One could only apply it for certain time steps in the reverse diffusion process, only apply it during inference on an unconditionally trained model \citep{choi_ilvr_2021} or using guidance, which applies a weighted sum of a conditional and unconditional generation and hence trades-off sample diversity with sample quality. Among others, guidance can be performed with an additionally trained classifier \citep{nichol_improved_2021}, by training the diffusion model conditionally and unconditionally at the same time and sampling using a weighted sum of both \citep{ho_classifier-free_2022} or by using pre-trained CLIP embeddings \citep{nichol_towards_2022,ramesh_hierarchical_2022}.

\subsection{Monocular Depth Estimation} Knowledge on the depth of a scene is crucial in a vast number of applications including virtual reality (VR), augmented reality (AR), environment perception, autonomous driving, robotics, state estimation, navigation, mapping and many more. Various surveys \citep{bhoi_monocular_2019,zhao_monocular_2020,ming_deep_2021, masoumian_monocular_2022} summarize the rich and extensive literature on monocular depth estimation; the estimation of the depth based on a single image; an inherently ill-posed and ambiguous problem.

In contrast, conventional geometric-based approaches such as structure from motion and stereo vision matching rely on geometric constraints and multiple viewpoints to reconstruct 3D structures. On the other hand, sensor-based methods leverage dedicated hardware like LiDAR sensors or RGB-D cameras to directly capture depth information. While these methods find practical application, they suffer from significant limitations including high hardware expenses, imprecise and sparse predictions, and limited accessibility for consumer use.

Many different representations can be deployed to obtain depth information: 2D depth maps (dense or sparse), 3D point cloud, 3D pseudo point cloud predicted from other modality (e.g. stereo camera), camera independent disparity maps, light fields, neural radiance fields (NERFs, \cite{mildenhall2021nerf}), 3D meshes, voxels and height maps to name a few.

Numerous deep learning based approaches for monocular depth estimation have been researched in recent years. 
\cite{lu_self-supervised_2022}, \cite{chen_self-supervised_2018} and \cite{zhang_unsupervised_2020} train their models using stereo images while inferring only single view images. The model either learns the correspondence between the two views and can reconstruct the other view or the model incorporates the knowledge of a stereo camera into its weights and hence strengthen its capability to extract meaningful features from a single image. \cite{yue_self-supervised_2022} and \cite{ zhao_joint_2022} apply a multi-task training objective by not only predicting depth but also other related tasks (e.g. normal vector prediction) that help the model to learn a more profound representation and understanding of the scene which also benefits the downstream task of depth estimation.
\cite{watson_temporal_2021} and \cite{bian_unsupervised_2021} use mono camera videos to estimate the depth.

Other deep learning-based approaches focus on multiple sensor modalities to estimate the depth of the scene. \cite{zhang_lidar-guided_2022} use LiDAR point clouds in combination with stereo images, \cite{eldesokey_confidence_2020} use monocular RGB images combined with sparse LiDAR point clouds, \cite{liu_pseudo-lidar_2020} input monocular RGB combined with a depth map and \cite{piao_dynamic_2021} combine a single RGB image with a focal stack.

In this work, we focus on monocular depth estimation using single RGB images as input and generating dense depth maps as output.
We observed that most model architectures are based on GANs, VAEs or similar approaches. We hence further review the usage of diffusion models in the field of depth estimation.

\subsection{Depth Diffusion}
We observe that very little work has been published on monocular depth estimation using diffusion models. We hypothesize that one of the major reasons is the necessity for fast sampling constraint by real-time applications like autonomous driving. We are certain that the community will find a way to further increase the sampling speed in a future and hence we see great potential in diffusion models for monocular depth estimation.

\cite{saxena_monocular_2023} used a diffusion model to perform monocular depth estimation on indoor and outdoor scenes introducing a preprocessing step to complete incomplete depth data and were even able to resolve depth ambiguity introduced from transparent surfaces.
\cite{duan_diffusiondepth_2023} use a latent diffusion model in combination with a Swin Transformer backbone \citep{liu_swin_2021} to first encode the depth into latent space and then perform the reverse diffusion in the latent space by iteratively applying their light weighted monocular conditioned denoising block. Finally, they apply a fully convolutional decoder on the diffused latent space to obtain the final depth prediction.
\cite{ranftl_towards_2020} propose methods to mix multiple depth datasets for robust training to mitigate the challenges of acquiring dense ground truth data from a variety of scenes. \cite{bhat_zoedepth_2023} proposes a method to combine relative depth and metric depth in a zero-shot manner, by pre-training on a large corpus of relative depth datasets and finetuning on metric depth.

Not as closely related but still applying diffusion models on 3D related data representations, we found \textbf{3D point cloud generation} conditioned on monocular images \citep{zhou_3d_2021}, conditioned on an encoded shape latent \citep{luo_diffusion_2021} and conditioned in CLIP-tokens \citep{nichol_point-e_2022}, \textbf{novel view synthesis} from a single view \citep{rockwell_pixelsynth_2021,watson_novel_2022}, for perpetual view generations for long camera trajectories where depth is predicted as an intermediate representation \citep{liu_infinite_2021} or combining text prompts for 2D generation with Neural Radiance Fields (NeRFs) \citep{poole_dreamfusion_2022}, \textbf{depth estimation} from multiple camera images at different viewpoints \citep{khan_differentiable_2021} and \textbf{depth-aware guidance} methods that guide the image generation process by its intermediate depth representation \citep{kim_dag_2023}.
\section{Background Denoising Diffusion Probabilistic Models} \label{sec:background}
In this section we provide the mathematical foundation on denoising diffusion probabilistic models including recent advances. We will not cover continuous time score-based models nor latent diffusion models.

As stated in the introduction, diffusion models are generative deep learning models. In general, generative models are likelihood-based models, meaning they learn a data distribution that, ideally, maximizes the likelihood that the learned distribution matches the real data distribution. Specifically, diffusion models are latent variable models that learn to generate data, by sampling from a latent space distribution in a Markovian process, usually a Gaussian distribution due to its convenient closed form representation. This process is the reverse diffusion process the model must learn. The forward diffusion process, also a Markovian process, is applied during training to learn the reverse diffusion process.

\subsection{Forward Diffusion Process} \label{sec:forward_diffusion_process}
In the forward diffusion process, Gaussian noise $\mathcal{N}$ is repeatedly added to a input data sample $x_0$ for $t$ time steps with $\{t \in \mathbb{R} | 1 \le t \le T\}$ in a Markovian process, such that $p\left(x_T\right)=\mathcal{N}\left(x_T;0,\boldsymbol{I}\right)$.
A sample $x_t$ only depends on a sample $x_{t-1}$ and a fixed noise schedule $\beta_t$ and is defined by the forward diffusion kernel:
\begin{equation}
\label{eq:diffusion_kernel}
q\left(x_t|x_{t-1}\right) = \mathcal{N}\left(x_t;\sqrt{1-\beta x_{t-1}},\beta_t \boldsymbol{I}\right)
\end{equation}

The joint distribution $q\left(x_{1:T}|x_{0}\right)$ of all samples generated on the trajectory of the Markovian forward diffusion process is defined as the product of the diffusion kernel at each time step $t$:
\begin{equation}
q\left(x_{1:T}|x_{0}\right) = \prod_{t=1}^{T} q\left(x_t|x_{t-1}\right)
\end{equation}

Equation \eqref{eq:diffusion_kernel} can be simplified to be able to generate a noisy sample at any given time step $t$ only conditioned on $x_0$:
\begin{equation}
\label{eq:diffusion_kernel_simplified}
\begin{aligned}
q\left(x_{t}|x_{0}\right) &= \mathcal{N}\left(x_t;\sqrt{\bar{\alpha}_t}x_0,\left(1-\bar{\alpha}_t \right) \boldsymbol{I}\right) \\
\bar{\alpha}_t &= \prod_{s=1}^{t} \left(1-\beta_t\right)
\end{aligned}
\end{equation}

The noise schedule $\beta_{t}$ is selected in such a way, that for $T$ time steps, $\bar{\alpha_T}$ approaches 0, which, according to \eqref{eq:diffusion_kernel_simplified}, results in a standard normal distribution $\mathcal{N}\left(0,\boldsymbol{I}\right)$ for $q\left(x_{T}|x_{0}\right)$. The shape of the noise schedule is a hyperparameter to be selected. While \cite{ho_denoising_2020} uses a linear schedule, \cite{nichol_improved_2021} proposes a cosine schedule.

To be able to calculate gradients of a stochastic variable in the back-propagation step during training, it is required to apply the reparameterization trick for sampling from a Gaussian distribution. A sample $x_t$ at a given time step $t$ can be formulated as:
\begin{equation}
x_{t} = \sqrt{\bar{\alpha}_t}x_0 + \sqrt{\left(1-\bar{\alpha}_t\right)}\epsilon, \;  where \: \epsilon \sim \mathcal{N}\left(0,\boldsymbol{I}\right)
\end{equation}

The data distribution $q\left(x_{t}\right)$ at any time step $t$ is the joint probability of all distributions of previous time steps. Using ancestral sampling it can be reformulated to:
\begin{equation}
q\left(x_{t}\right) = \int q\left(x_{0},x_{t}\right) dx_0 = \int q\left(x_{0}\right) q\left(x_{t}|x_{0}\right) dx_0
\end{equation}
In other words, to draw a sample $x_t \sim q\left(x_{t}\right)$, one can first draw $x_0 \sim q\left(x_{0}\right)$, which is basically drawing a sample from the input dataset, and then draw a sample $x_t$ from $q\left(x_{t}|x_0\right)$, which is the forward diffusion from equation \eqref{eq:diffusion_kernel_simplified}.

\subsection{Reverse Diffusion Process} \label{sec:reverse_diffusion_process}

For DDPMs, the reverse diffusion process is Markovian, similar to that of the forward diffusion process. 
The naive approach is to draw the initial sample $x_T \sim \mathcal{N}\left(x_T;\boldsymbol{0},\boldsymbol{I}\right)$ and then iteratively draw a less noisy sample $x_{t-1} \sim q\left(x_{t-1}|x_{t}\right)$. The issue is that $q\left(x_{t-1}|x_{t}\right)$ is intractable, meaning there is no closed form solution. Diffusion models solve this issue by learning the intractable posterior distribution parameterized by $\theta$ given as:
\begin{equation}
\label{eq:intractable_posterior_distribution}
p_{\theta}\left(x_{t-1}|x_t\right) = \mathcal{N}\left(x_{t-1};\mu_{\theta}\left(x_t,t\right),\Sigma_{\theta}\left(x_t,t\right)\right)
\end{equation}

and its corresponding joint distribution
\begin{equation}
\label{eq:joint_reverse_distribution}
p_{\theta}\left(x_{1:T}|x_{0}\right) = p\left(x_{T}\right)\prod_{t=1}^{T} p_{\theta}\left(x_{t-1}|x_{t}\right)
\end{equation}

The diffusion model is trained by optimizing the variational bound of the negative log-likelihood: 
\begin{equation}
\begin{aligned}
L & := \mathbb{E}_{q\left(x_{0}\right)}\left[-log\, p_{\theta}\left(x_{0}\right)\right] \\
  & \le \mathbb{E}_{q\left(x_{0}\right)q\left(x_{1:T}|x_0\right)}\left[-log\frac{p_{\theta}\left(x_{0:T}\right)}{q\left(x_{1:T}|x_0\right)}\right]
\end{aligned}
\end{equation}

which can be rewritten as
\begin{equation}
\label{eq:variational_lower_bound}
\begin{aligned}
L_{VLB} & := \mathbb{E}_{q}\left[L_{0} + L_{t-1} + L_{T}\right] \\
L_{0} & := -log\, p_{\theta}\left(x_{0}|x_{1}\right)\\
L_{t-1} & := \sum_{t<1}D_{KL}\left(q\left(x_{t-1}|x_t,x_0\right)\|p_{\theta}\left(x_{t-1}|x_t\right)\right)\\
L_{T} & := D_{KL}\left(q\left(x_{T}|x_0\right)\|p\left(x_{T}\right)\right)
\end{aligned}
\end{equation}
where $D_{KL}\left(\|\right)$ is the Kullback-Leibler(KL)-Divergence between two distributions and $q\left(x_{t-1}|x_t,x_0\right)$ is the tractable posterior distribution
\begin{equation}
q\left(x_{t-1}|x_t,x_0\right) = \mathcal{N}\left(x_{t-1};\Tilde{\mu}_t\left(x_t,x_0\right),\Tilde{\beta}_t\boldsymbol{I}\right)
\end{equation}
where
\begin{equation}
\begin{aligned}
\Tilde{\mu}_t\left(x_t,x_0\right) & := \frac{\sqrt{\bar{\alpha}_{t-1}}\beta_t}{1-\bar{\alpha}_{t}}x_0 + \frac{\sqrt{1-\beta_t}\left(1-\bar{\alpha}_{t-1}\right)}{1-\bar{\alpha}_{t}}x_t \\
\Tilde{\beta}_t & := \frac{1-\bar{\alpha}_{t-1}}{1-\bar{\alpha}_{t}}\beta_t
\end{aligned}
\end{equation}

In other words, $\Tilde{\mu}_t\left(x_t,x_0\right)$ is the weighted sum of an unnoisy sample $x_0$ and a noisy sample $x_t$ at time step $t$.

Equation \eqref{eq:variational_lower_bound} can be further simplified by setting $L_T=0$, since the Kullback-Leibler-Divergence of two Gaussians is zero under the assumption that $q\left(x_{T}|x_0\right) \approx \mathcal{N}\left(\boldsymbol{0},\boldsymbol{I}\right)$ and $p\left(x_{T}\right) \approx \mathcal{N}\left(\boldsymbol{0},\boldsymbol{I}\right)$, which previously has been defined to be the case.

The better the model learns to approximate the parameterized denoising posterior distribution $p_{\theta}\left(x_{t-1}|x_t\right)$ with the real tractable denoising posterior distribution $q\left(x_{t-1}|x_t,x_0\right)$, the smaller the KL-divergence and hence the smaller $L_{VLB}$.
Since all distributions in \eqref{eq:variational_lower_bound} are tractable and all KL-Divergences are comparisons of Gaussians, they can be calculated in closed form:
\begin{equation}
\label{eq:closed_form_lt-1}
\begin{aligned}
L_{t-1} & = D_{KL}\left(q\left(x_{T}|x_0\right)\|p\left(x_{T}\right)\right) \\
        & = \mathbb{E}_{q}\left[\frac{1}{2\sigma^2_t}\|\Tilde{\mu}_t\left(x_t,x_0\right)-\Tilde{\mu}_{\theta}\left(x_t,t\right)\|^2\right] + const
\end{aligned}
\end{equation}

\cite{ho_denoising_2020} found that predicting the noise $\epsilon$ that was applied in the forward diffusion to reverse the diffusion process by using a noise predictor $\epsilon_{\theta}$ works best and modified \eqref{eq:closed_form_lt-1} to:
\begin{equation}
\begin{aligned}
L_{t-1} = \mathbb{E}_{q}\left[\lambda_t\frac{\beta_t}{\left(1-\beta_t\right)\left(1-\alpha_t\right)}\|\epsilon-\epsilon_{\theta}\left(x_t,t\right)\|^2\right] + const
\end{aligned}
\end{equation}

\cite{ho_denoising_2020} further observed that setting the time dependent scalar $\lambda_t=\left(1-\beta_t\right)\left(1-\alpha_t\right)/\beta_t$, improves sample quality and simplifies the training objective to:
\begin{equation}
\label{eq:loss_simple}
\begin{aligned}
L_{simple} = \mathbb{E}_{q}\left[\|\epsilon-\epsilon_{\theta}\left(x_t,t\right)\|^2\right].
\end{aligned}
\end{equation}

In contrast to \cite{ho_denoising_2020}, \cite{choi_perception_2022} proposes a more sophisticated choice of $\lambda_t$, namely P2 weighting, to prioritizes different noise levels to improve the sample quality:
\begin{equation}
\lambda'_t = \frac{\lambda_t}{\left(k+SNR(t)\right)^{\gamma}}
\end{equation}
where $k$ is a hyper parameter to prevent exploding weights, $\gamma$ controls the strength of down weighting and the signal-to-noise ratio $SNR(t)=\alpha_t/(1-\alpha_t)$ is a simplified expression for the noise schedule by \cite{kingma_variational_2022}.

While \cite{ho_denoising_2020} only predicts the mean of the added noise and sets the variance to be either $\beta_t$ or $\Tilde{\beta}_t$ the upper and lower variational bound respectively, \cite{nichol_improved_2021} found that learning the variance $\Sigma_{\theta}\left(x_t,t\right)$ from equation \eqref{eq:intractable_posterior_distribution} improves sample quality and allows sampling with a reduced number of time steps with little change in sample quality. Instead of predicting $\Sigma_{\theta}\left(x_t,t\right)$ directly, they propose to learn the variance as a weighted sum of the upper and lower bound using a neuronal network's output $v$:
\begin{equation}
\Sigma_{\theta}\left(x_t,t\right) = exp\left(v\,log\,\beta_t+\left(1-v\right)log\,\beta_t\right)
\end{equation}

Since $L_{simple}$ does not provide a learning signal for $\Sigma_{\theta}\left(x_t,t\right)$, \cite{nichol_improved_2021} proposes a hybrid loss function:
\begin{equation}
L_{hybrid} = L_{simple} + \lambda L_{VLB}
\end{equation}

To sample a less noisy sample $x_{t-1}\sim p_{\theta}\left(x_{t-1}|x_t\right)$ the trained diffusion model $\epsilon_{\theta}$ estimates the noise that was added from time step $t-1$ to $t$ and subtracts that noise from $x_t$.
\begin{equation}
x_{t-1} = \frac{1}{\sqrt{\alpha_t}}\left(x_t - \frac{1-\alpha_t}{\sqrt{1-\bar{\alpha}_t}}\epsilon_{\theta}(x_t,t)\right) + \sigma_t z
\end{equation}
where $ z \sim \mathcal{N}\left(\boldsymbol{0},\boldsymbol{I}\right) $

\subsection{Conditional Diffusion Process} \label{sec:conditional_diffusion_process}
All previous formulations describe the unconditional case, where unconditional means no extra condition beside the Markovian process. The ultimate goal of a generative model is to control the sampling process by incorporating a condition $c$ to obtain a desired output.

The reverse process from equations \eqref{eq:intractable_posterior_distribution} and \eqref{eq:joint_reverse_distribution} can be extended as follows:
\begin{equation}
\begin{aligned}
p_{\theta}\left(x_{t-1}|x_t, c\right) & = \mathcal{N}\left(x_{t-1};\mu_{\theta}\left(x_t,t,c\right),\Sigma_{\theta}\left(x_t,t,c\right)\right) \\
p_{\theta}\left(x_{0:T}|c\right) & = p\left(x_{T}\right)\prod_{t=1}^{T} p_{\theta}\left(x_{t-1}|x_{t},c\right)
\end{aligned}
\end{equation}

Similarly, the variational lower bound from equation \eqref{eq:variational_lower_bound} is extends to:
\begin{equation}
\label{eq:variational_lower_bound_conditional}
\begin{aligned}
L_{VLB} & := \mathbb{E}_{q}\left[L_{0} + L_{t-1} + L_{T}\right] \\
L_{0} & := -log\, p_{\theta}\left(x_{0}|x_{1},c\right)\\
L_{t-1} & := \sum_{t<1}D_{KL}\left(q\left(x_{t-1}|x_t,x_0\right)\|p_{\theta}\left(x_{t-1}|x_t,c\right)\right)\\
L_{T} & := D_{KL}\left(q\left(x_{T}|x_0,c\right)\|p\left(x_{T}\right)\right)
\end{aligned}
\end{equation}
\section{Proposed Method}\label{sec:proposed_method}
In this section we introduce our proposed method. We first show our framework \modelname{} and then provide further details on the architectural composition.

\subsection{Framework} \label{subsec:Framework}

\begin{figure}[t]
  \centering
  \includegraphics[width=0.99\linewidth]{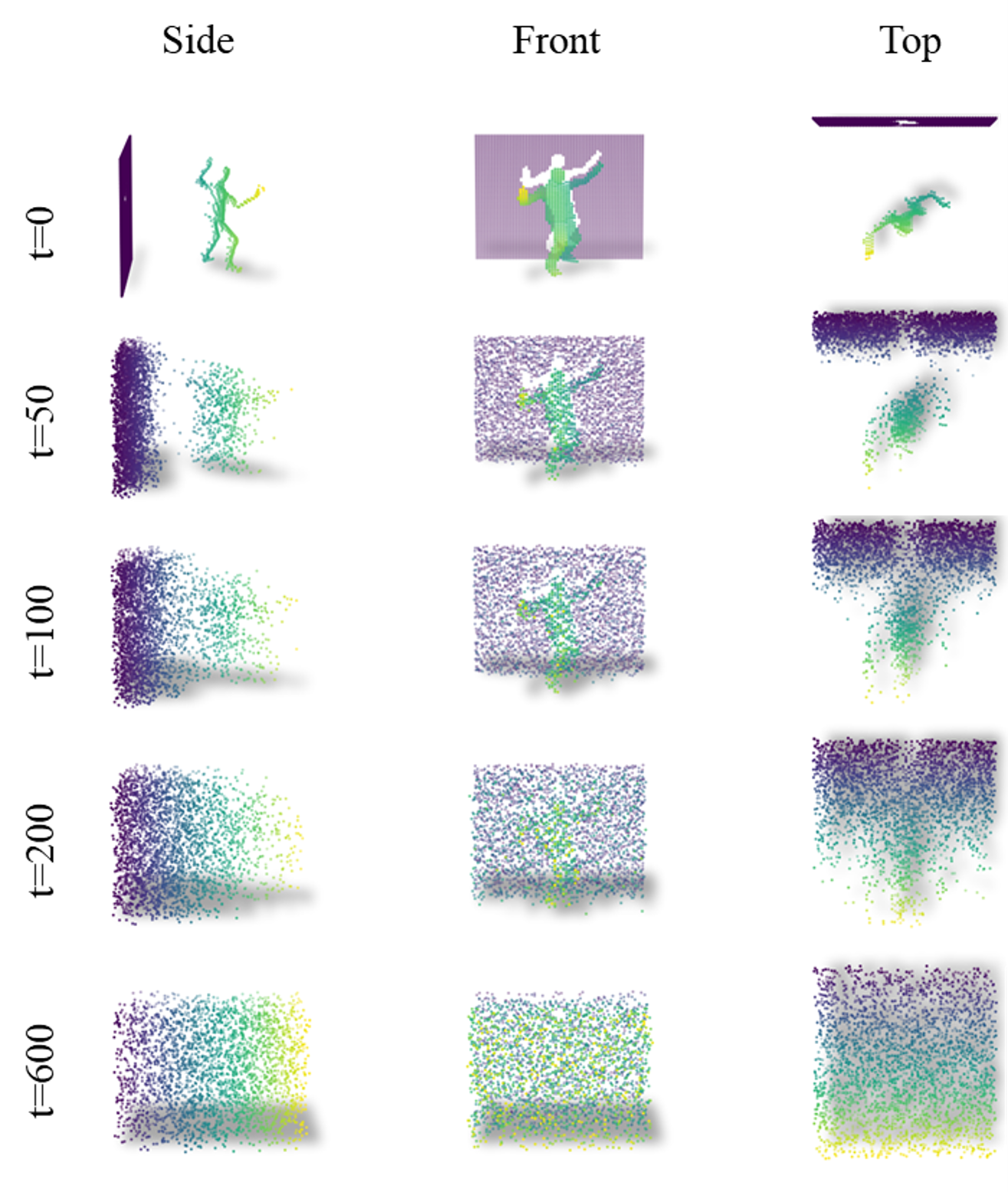}
  \caption{Forward diffusion process of a depth map represented as point cloud $P_{d}$ with a cosine beta schedule and with T=600 diffusion steps. Depth values are scaled to range from -1 to 1, with -1 being the background.}
  \label{fig:forward_diffusion}
\end{figure}

\begin{figure*}[t]
  \centering
  \includegraphics[width=0.99\textwidth]{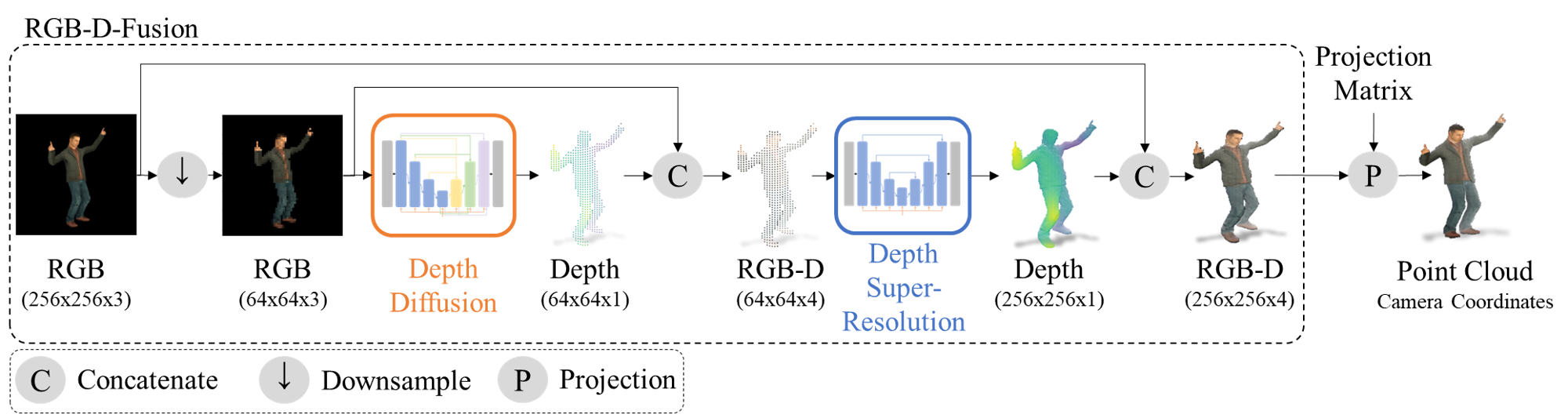}
  \caption{\modelname{} framework. Our framework takes an RGB image as input and outputs an RGB-D image using perspective projection. We first downsample the input image from 256x256 to 64x64. We then predict the perspective depth map using a conditional denoising diffusion probabilistic model. We then combine the predicted depth map with the input RGB into an RGB-D image of resolution 64x64. We further apply a super-resolution model conditioned on the low-resolution RGB-D to obtain a high-resolution depth map. The predicted high-resolution depth map is combined with the input RGB to construct the final output: a high-resolution RGB-D image. To reconstruct the real depth in camera coordinates, a projection matrix can be applied if available.}
  \label{fig:framework}
\end{figure*}

The \modelname{} framework is depicted in Fig. \ref{fig:framework}. We input an RGB image of a person with removed background (e.g. captured from a consumer camera device) and output an RGB-D image of that same person, with perspective projection. 

Our framework consists of two stages: a depth prediction stage and a depth super-resolution stage. 
The first stage outputs a low-resolution perspective depth map for a given low-resolution RGB image. The second stage outputs a high-resolution depth map for a given low-resolution RGB-D input.

We first downsample the RGB input to a lower resolution of 64x64 pixels. We resample RGB data using bilinear resampling while for depth maps we apply nearest neighbors resampling to avoid undesired artefacts due to the large gradients introduced of the removed background. We feed the downsampled image into our first stage, a conditional diffusion model (details in section \ref{subsubsec:rgb_conditioned_depth_diffusion_model}) to predict the corresponding depth map. We combine the low-resolution RGB input with the predicted depth map to form an RGB-D image. The low-resolution RGB-D image is fed into our second stage, another conditional diffusion model (details in section \ref{subsubsec:rgbd_conditioned_depth_superresolution_diffusion_model}) to predict a high-resolution depth map. Finally, the predicted depth is combined with the initial RGB input to form the final RGB-D output.

Both our models, the depth diffusion model and the depth super-resolution model, perform the diffusion process on a depth map. The depth diffusion model samples a low-resolution depth map in 64x64 pixels, conditioned on a low-resolution RGB image. Our depth super-resolution model is conditioned on a low-resolution RGB-D image and samples a high-resolution depth map in 256x256 pixels. To obtain the depth in camera coordinates, one must further project the predicted output using a projection matrix if available.

We a apply a cosine beta schedule with $T=600$ for our depth diffusion model and $T=1000$ for our depth super-resolution model. Fig. \ref{fig:forward_diffusion} depicts the forward diffusion of a depth map represented as point cloud for various time steps and viewpoints.

In section \ref{sec:experiments} we perform various experiments to ablate and justify our architectural decision now introduced in further detail in section \ref{subsec:model_architecture}.

\subsection{Dataset}\label{subsec:dataset}

We created a custom dataset by rendering RGB-D images using perspective projection from high-quality 3D models of people \citep{volograms2021,escribano2022texture}. We randomly varied the camera's viewpoint, its field of view and its distance to the subject in such a way that the projected person covers 80\% of the image's height. Our dataset contains $\approx$ 25k samples. Each sample contains an RGB image of a person and its corresponding depth map. Both modalities have no background and have a spatial resolution of 256x256. 
The $\approx$ 25k samples are composed of 358 different subjects. We divide the dataset into a train set with 19,902 samples containing 315 subjects and a test set with 5,302 samples of 43 subjects. For better visualization of the depth maps, we illustrate them as point cloud $P_d = \{\left(u_{i}, v_{i},z_{i}\right) | i \in N \}$, where $N$ is the number of pixels of the depth map, $u_{i}$ and $v_{i}$ are the pixel coordinates and $d_{i}$ is the depth value. Likewise, an RGB-D image is represented as point cloud $P_{rgbd}= \{\left(u_{i}, v_{i}, d_{i},r_{i}, g_{i}, b_{i}\right) | i \in N \}$. In contrast, a colored point cloud in camera coordinates $P_{C}= \{\left(x_{i}, y_{i}, z_{i},r_{i}, g_{i}, b_{i}\right) | i \in N \}$ is defined by its 3D coordinates $x_i$, $y_i$ and $z_i$.

\subsection{Model Architecture} \label{subsec:model_architecture}
In this section we provide details on the two models that form our \modelname{} framework: the RGB conditioned depth diffusion model and the RGB-D conditioned depth super-resolution diffusion model.

\subsubsection{Base Model} \label{subsubsec:base_model}
Before we introduce the individual models we show the base model architecture from which we later derive our depth diffusion model and our depth super-resolution model.

\begin{figure*}[t]
  \centering
  \includegraphics[width=0.99\textwidth]{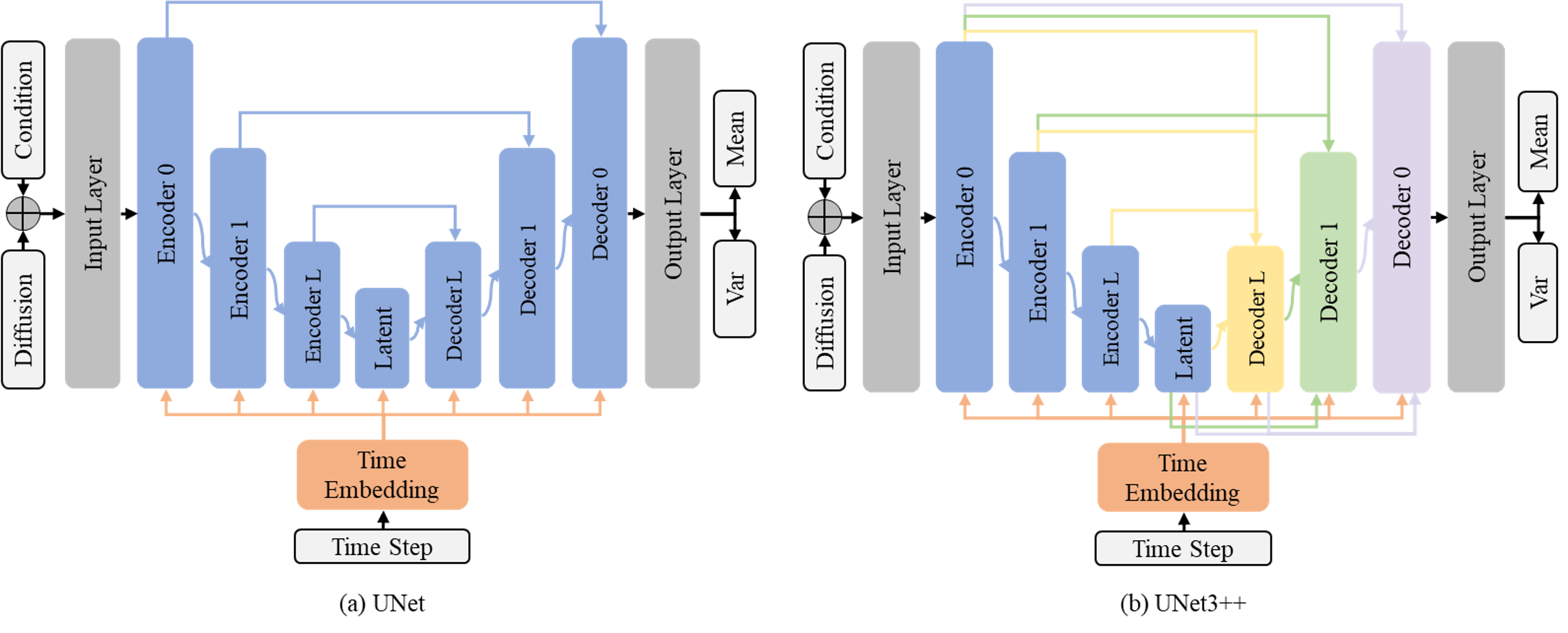}
  \caption{Comparison of our two base architectures. We condition our models on the time step $t$ of the diffusion process and on an additional condition input that is concatenated with the diffusion input. The diffusion input is the noisy sample. During training noise is gradually added to the diffusion input and gradually removed during sampling, while the condition is kept unchanged. Our models predict the mean and the variance of the conditioned diffusion input. The standard UNet architecture depicted in (a) simply connects the encoder and decoder stages with the same spatial resolution, while for the UNet3+ in (b), the decoder is connected to all encoder stages with the same or higher spatial resolution and all decoders with a smaller spatial resolution. Before concatenation, all feature maps are resampled to the same spatial resolution.}
  \label{fig:base_architecture}
\end{figure*}

In contrast to previous works \citep{dhariwal_diffusion_2021,nichol_improved_2021, ho_cascaded_2021} our base model architecture for the depth diffusion model is based on UNet3+ \citep{huang_unet_2020} and the architecture for our super-resolution model is based on the UNet architecture. A high level comparison is shown in Fig. \ref{fig:base_architecture}.

Similar to \citep{dhariwal_diffusion_2021,nichol_improved_2021} we implement the stages of the model using multiple ResNet blocks followed by an optional multi-head attention block with 8 heads and a hidden dimension of 32 per head. For our latent space stage we implement standard attention by \cite{vaswani_attention_2017} and for stages with higher resolution we implement linear attention as proposed by \cite{wang_linformer_2020}. We also implement residual up and down-sampling blocks as suggested in BigGAN by \cite{brock_large_2019}.
We replace all normalization layers (i.e. batch normalization \cite{ioffe_batch_2015} and instance normalization \citep{ulyanov_instance_2017}) with group normalization \citep{wu_group_2018} with a groupsize of 8 and $\epsilon = 1e-05$ and only keep layer normalization \citep{ba_layer_2016} for all our attention blocks. 
As suggested by \cite{ioffe_batch_2015}, we remove all bias weights from our convolutional layers since the normalization layers introduce a learnable bias.
All activations are GeLU \citep{hendrycks_gaussian_2020} except for the softmax activations in the attention blocks.
We input the condition via concatenation with the noise input.
The time step information is input into each ResNet block (standard as well as in up and down sampling) using adaptive group normalization (AdaGN) \citep{dhariwal_diffusion_2021}.
Following \cite{liu_convnet_2022}, we increase the number of consecutive ResNet blocks in lower resolution stages and further apply stochastic depth \citep{huang_deep_2016} during training, to stochastically drop entire ResNet blocks to train an implicit ensemble of models.

\subsubsection{RGB Conditioned Depth Diffusion Model} \label{subsubsec:rgb_conditioned_depth_diffusion_model}
Our depth diffusion model is based on the UNet3+ architecture conditioned on a low-resolution (64x64x3) RGB image and generates a low-resolution depth map (64x64x1) when sampled from. 
The intention behind this architectural choice is to strengthen the decoder by using inputs from multiple resolutions of the UNet encoder, hence combining local and global features.
Corresponding experiments are performed in section \ref{subsec:experiements_depth_diffusion}.

\subsubsection{RGB-D Conditioned Depth Super-Resolution Diffusion Model} \label{subsubsec:rgbd_conditioned_depth_superresolution_diffusion_model}

\begin{figure}[t]
  \centering
  \includegraphics[width=0.99\linewidth]{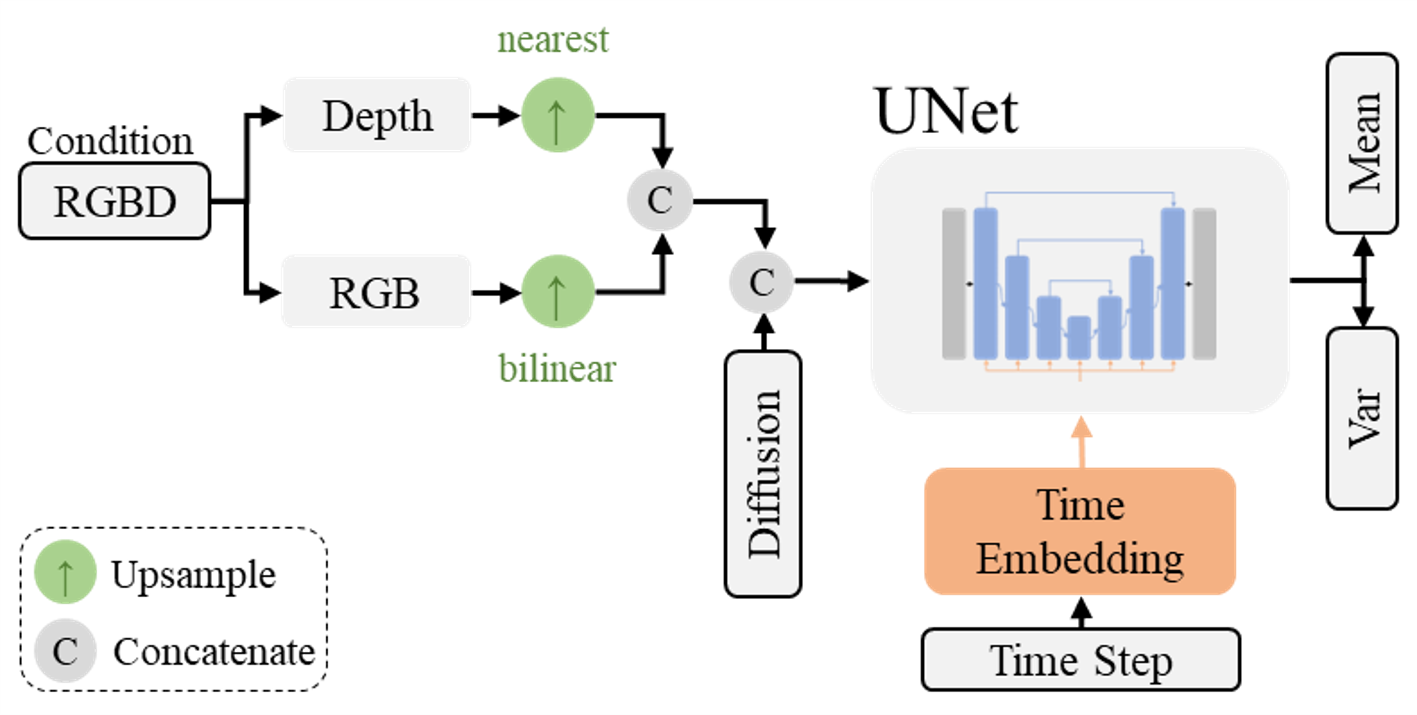}
  \caption{Architecture of our depth super-resolution model. The low-resolution RGB-D image condition is split into its depth and RGB components, upsampled using nearest neighbor and bilinear resampling respectively, and combined again before it is fed into a UNet model.}
  \label{fig:superres_model}
\end{figure}

Our super-resolution diffusion model is conditioned on a low-resolution (64x64x4) RGB-D image and generates a high-resolution depth map (256x256x1) when sampled from. We use a UNet base architecture as depicted in Fig. \ref{fig:base_architecture}. The only difference is that before the low-resolution condition is fed into the model, we first upsample to the target resolution to match the resolution of the diffusion input. The RGB-D input is split into RGB and depth map, we then apply bilinear resampling for the RGB image and nearest neighbor resampling for the depth map and finally concatenate to a high-resolution RGB-D image (See Fig. \ref{fig:superres_model}). 
Corresponding experiments are performed in section \ref{subsec:experiements_superresolution}.

Compared to our depth diffusion model we make the following two assumptions: first, we assume that super-resolution is an easier task compared to depth estimation conditioned on a different modality, hence less parameters are required to learn this task. Second, high-resolution stages in the UNet architecture are more important than low-resolution stages since it is more important to focus on local features and not to have a global understanding of the image's content. 
Based on these assumptions and the fact that higher resolution images require more resources for training, we made the following design changes compared to our depth estimation network: first, we use a the UNet as a base architecture and increase the number of channels in the first stages of the UNet. This gives more importance on high-resolution stages and decreases the importance for low-resolution stages, while at the same time reducing the number of parameters. To extract more features for a given resolution, we increase the number of blocks within each stage. Second, we use the same number of stages in the UNet for our super-resolution from 64 to 128 as for 64 to 256, following the same argumentation of higher importance of high-resolution stages. The reduced number of stages, reduces the number of trainable parameters and computations required for the forward pass and lets us apply a higher batch size, hence speeds up the training and inference time.

\subsection{Data Augmentation} \label{subsec:data_augmentation}

During the training of our depth diffusion model we randomly scale and shift the RGB condition input.

For our depth super-resolution model we follow the suggestion of \cite{ho_cascaded_2021} and augment the RGB condition input with a Gaussian blur by convolving the image with a Gaussian 3x3 kernel with a standard deviation of $\sigma_{rgb} \sim \mathcal{U}_{[0,0.6]}$ with a probability of 0.5 in each epoch:
\begin{equation}
\label{eq:rgb_aug}
    rgb_{aug} = rgb * \epsilon_{rgb}
\end{equation}
where $\epsilon_{rgb} \sim \mathcal{N}\left(0,\sigma_{rgb}\right)$, $\epsilon_{rgb} \in \mathbb{R}^{3x3x3}$ and $rgb,rgb_{aug} \in \mathbb{R}^{HxWx3}$

In contrast to \cite{ho_cascaded_2021}, we further provide the depth as a condition input to our super-resolution model. Following the thoughts of \cite{ho_cascaded_2021} we augment the depth $d$ by adding Gaussian noise with an empirically determined standard deviation of $\sigma_{d} \sim \mathcal{U}_{[0,0.06]}$:
\begin{equation}
\label{eq:depth_aug}
    d_{aug} = d + \epsilon_{d}
\end{equation}
where $\epsilon_{d} \sim \mathcal{N}\left(0,\sigma_{d}\right)$ and $\epsilon_{d},d,d_{aug} \in \mathbb{R}^{HxWx1}$

\begin{figure}[t]
  \centering
  \includegraphics[width=0.99\linewidth]{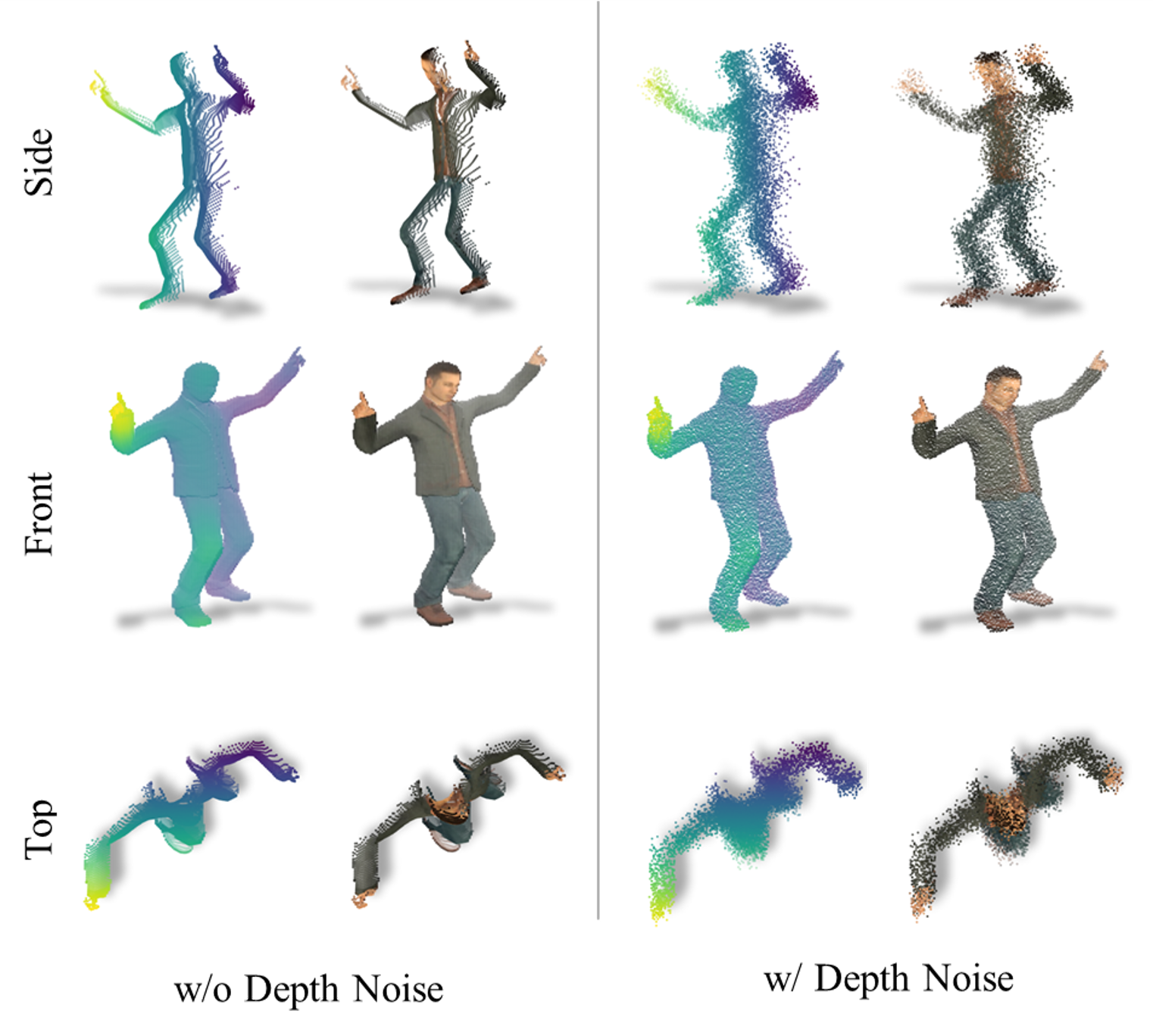}
  \caption{Comparison of a depth sample and an RGB-D sample without depth noise augmentation and with depth noise augmentation represented as point cloud $P_{d}$ and $P_{rgbd}$. We remove the points associated with the background for better representation.}
  \label{fig:depth_gaussian_augmentation}
\end{figure}

Fig. \ref{fig:depth_gaussian_augmentation} shows the comparison of the depth without and with Gaussian additive noise, represented as a point cloud.

We further apply random scaling and shifting of the RGB-D condition input.
\section{Experiments} \label{sec:experiments}

\begin{table*}[t]
\centering
\caption{Results of experiments performed with our depth diffusion model. We alter the model architecture, the method to obtain the variance, the loss formulation, the learning rate schedule, the number of residual blocks per stage and the stochastic depth probability. Underlined parameters depict the change compared to the previous run. The detailed architecture of the models can be found in TABLE \ref{tab:experiment_model_description_depth_diffusion} in the appendix.}
\label{tab:depth_diffusion_ablation}
\begin{tabular}{ l | c c c c c c | r | r | r | r }
\hline
\textbf{Run} & \textbf{Model} & \textbf{Variance} & \textbf{Loss} & \textbf{$L_r$ Schedule} & \textbf{ResBlocks} & \textbf{Stochastic Depth} & \textbf{MAE $\downarrow$} & \textbf{MSE} $\downarrow$ & \textbf{IoU} $\uparrow$  & \textbf{VLB} $\downarrow$ \\ 
\hline
\hline
 dd1 & UNet & fix & simple & \textbf{x} & 2/2/2/2 & 0/0/0/0 & 8.14 & 2.57 & 0.987 & 10.03 \\
 \hline
 dd2 & \underline{UNet3+} & fix & simple & \textbf{x} & 2/2/2/2 & 0/0/0/0 & 9.24 & 2.97 & 0.988 & \textbf{9.72} \\
 dd3 & UNet3+ & \underline{learned} & simple & \textbf{x} & 2/2/2/2 & 0/0/0/0 & 10.83 & 3.21 & 0.987 & 13.00 \\ 
 dd4 & UNet3+ & learned & \underline{P2} & \textbf{x} & 2/2/2/2 & 0/0/0/0 & 8.36 & 1.75 & 0.989 & 14.26 \\ 
 dd5 & UNet3+ & learned & P2 & \underline{\checkmark} & 2/2/2/2 & 0/0/0/0 & 7.70 & 1.58 & \textbf{0.993} & 12.20 \\ 
 dd6 & \underline{UNet} & learned & P2 & \checkmark & 2/2/2/2 & 0/0/0/0 & 7.45 & \textbf{1.48} & 0.990 & 12.39 \\
 dd7 & UNet & learned & \underline{simple} & \checkmark & 2/2/2/2 & 0/0/0/0 & 15.65 & 8.75 & 0.946 & 20.48 \\ 
 dd8 & \underline{UNet3+} & learned & simple & \checkmark & 2/2/2/2 & 0/0/0/0 & 6.16 & 1.64 & \textbf{0.993}  & 17.59 \\
 dd9 & UNet3+ & learned & simple & \checkmark & \underline{2/2/12/2} & 0/0/0/0 & 231.91 & 251.30 & 0.539 & 23.97 \\
 dd10 & UNet3+ & learned & simple & \checkmark & 2/2/12/2 & \underline{0.1/0.1/0.5/0.1} & \textbf{5.79} & \textbf{1.48} & \textbf{0.993} & 16.95  \\
 \hline
\end{tabular}
\end{table*}

\begin{figure*}[t]
  \centering
  \includegraphics[width=0.99\textwidth]{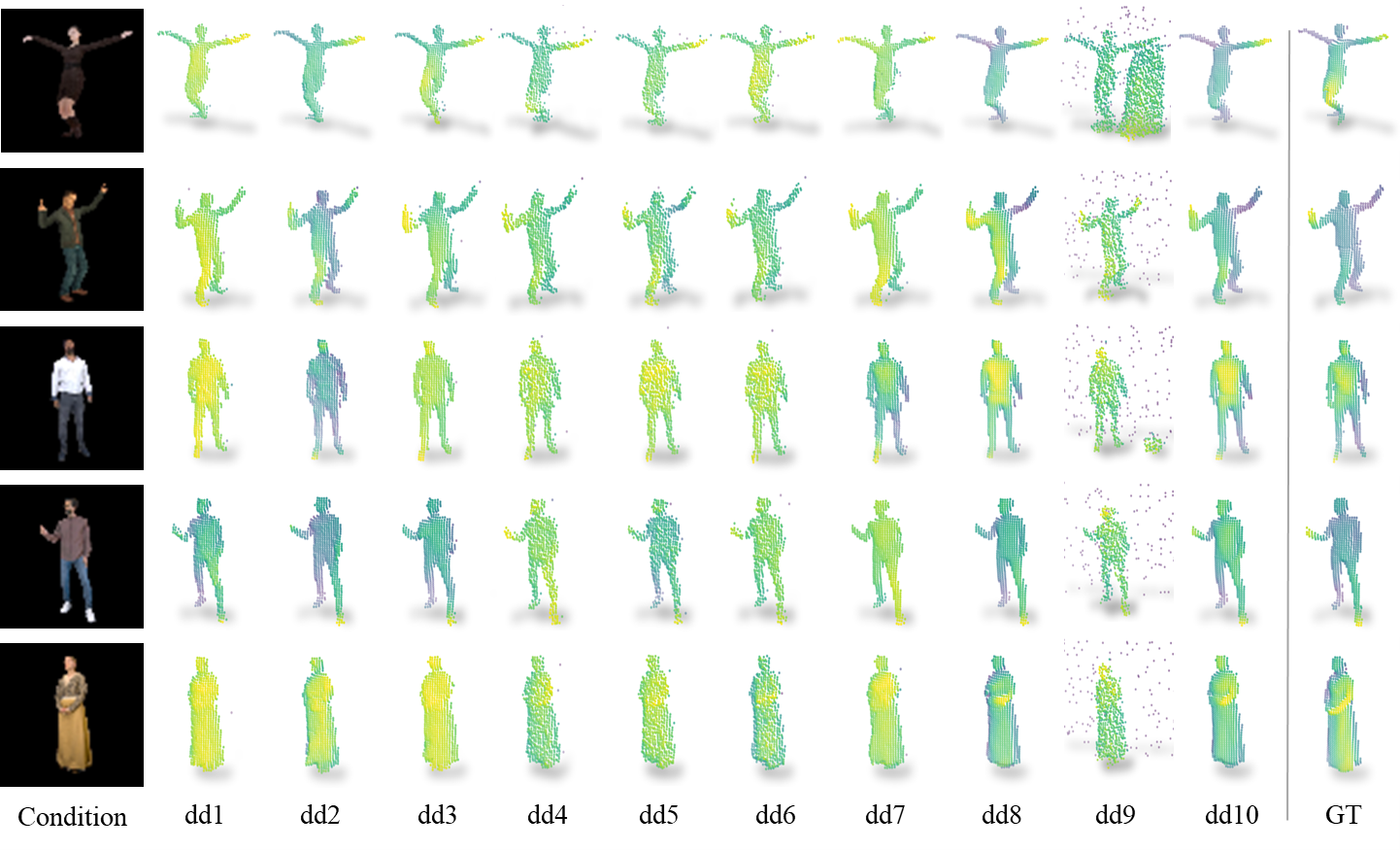}
  \caption{Generated depth maps represented as point clouds $P_{d}$ from our depth diffusion models evaluated in our experiments from TABLE \ref{tab:depth_diffusion_ablation} for each run respectively. To better observe the details, the point clouds' viewpoint has been rotated 10 degrees in azimuth and 10 degrees in elevation. The color map encodes the depth between the min and max value of the depth of each plot, where darker colors indicate higher depth and brighter colors closer depth.}
  \label{fig:depth_ablation}
\end{figure*}

In this section we perform experiments to validate our design choices. We first run multiple experiments on the conditional depth diffusion network to select various hyperparameters related to model architecture and model training. We incorporate our findings into the conditional depth super-resolution model and find that the best architecture for the depth diffusion is not suitable in terms of resources and performance on the metrics. We hence modify the architecture, investigate on which input modality to condition to and perform an ablation study on the augmentation of the condition input. 
We evaluate our experiments with the following metrics: 
\begin{itemize}
  \item \textbf{MAE} mean average error distance of the ground truth depth $d_{gt}$ and the predicted depth $d_{pred}$ with $d_{gt}, d_{pred} \in \mathbb{R}^{HxWx1}$.
  \begin{equation}
      MAE = \frac{1}{n}\sum_{i=1}^{n}\left|d_{gt}-d_{pred}\right| \cdot 10^{3}
  \end{equation}
  We multiply the result with $10^{3}$. The closer to zero the better.
  \item \textbf{MSE} mean squared error distance of the ground truth depth $d_{gt}$ and the predicted depth $d_{pred}$.
  \begin{equation}
    MSE = \frac{1}{n}\sum_{i=1}^{n}\left(d_{gt}-d_{pred}\right)^2 \cdot 10^{3}
  \end{equation}
  We multiply the result with $10^{3}$. The closer to zero the better.
  \item \textbf{IoU} Intersection over Union calculated over a binary mask $m$ of the ground truth $gt$ and the prediction $p$.  
  \begin{equation}
    IoU = \frac{m_{gt} \cap m_{p}}{m_{gt} \cup m_{p}}, with \; m_{y} = \left\{\begin{array}{ll} 1, & y > \phi \\
         0, & y \le \phi \end{array}\right. 
  \end{equation}
  where $\phi$ is the threshold for the binary mask. We set $\phi=-0.95$ since the depth maps range from [-1,1].
  The closer to 1 the better.
  \item \textbf{VLB} of the negative log-likelihood of the test data set (see \eqref{eq:variational_lower_bound}) reported in bits/dim with base-2 logarithm.
  The smaller the better.
\end{itemize}

As highlighted by \cite{theis_note_2016} the evaluation metrics of generative models are complex and might be misleading especially for high dimensional data like images. High log-likelihood does not imply high sample quality and low log-likelihood does not imply poor samples. Furthermore, an evaluation on samples is claimed to be biased towards overfitting models and favors models of large entropy. We therefore do not rely on a single metric and report the metrics listed above and show samples drawn from the model distribution. 

We further want to highlight the weakness of distance metrics like MSE and MAE since the pixel location is of high importance and even a slight shift of a few pixels of two indistinguishable images leads to a high distance metric, as reported by \cite{theis_note_2016} for a nearest neighbor distance.

All experiments are performed on our test set described in section \ref{subsec:dataset} which contains $\approx 5300$ data samples. 

\subsection{Depth Diffusion Model} \label{subsec:experiements_depth_diffusion}
We first investigate the impact of different hyper-parameters on mentioned metrics for our conditional depth diffusion model. Explicitly we perform 10 runs in which we vary the model architecture (UNet vs. UNet3+), the approach for determining the variance (learned range vs. fixed to upper bound $\beta_t$), the loss weighting approach (simple vs. P2), a learning rate schedule (no schedule vs. cosine decay with restart and linear warm-up), the number of residual blocks and the usage of stochastic depth augmentation. A detailed summary of the most important hyper parameters for all runs is listed in TABLE \ref{tab:experiment_model_description_depth_diffusion} in the appendix.

The results for the conditional depth diffusion models are summarized in TABLE \ref{tab:depth_diffusion_ablation} and outputs of the respective models are depicted in Fig. \ref{fig:depth_ablation}. The underlined parameter highlights the changed configuration compared to the previous run. Best results are printed in bold.

Our baseline (Run 1 in TABLE \ref{tab:depth_diffusion_ablation}) is a UNet model (similar to \cite{dhariwal_diffusion_2021}). The variance is not learned but set to $\beta_t$, the upper bound of the reverse diffusion process. We apply the simple loss $L_{simple}$ from equation \eqref{eq:loss_simple} as our training objective and apply no learning rate schedule. 

All runs are trained for 250 epochs using ADAM optimizer with a learning rate $\alpha$ of 4e-5 (exception of Run 5 with $\alpha=6e-5$), $\beta_1=0.9$ and $\beta_2=0.999$. The forward and the reverse diffusion process is performed with $T=600$ time steps and a cosine noise schedule as in \cite{nichol_improved_2021}. We further apply random scaling and shifting augmentation on the RGB input condition. 

Based on the results in TABLE \ref{tab:depth_diffusion_ablation} and the generated images in Fig. \ref{fig:depth_ablation} we consider run dd10 best.

\subsection{Depth Super-Resolution Model} \label{subsec:experiements_superresolution}
We perform various experiments for our super-resolution model. We compare different architectures and ablate the augmentation method. A detailed summary of the most important hyperparameters for all runs is listed in TABLE \ref{tab:experiment_model_description_superres_condition} in the appendix.

First, we investigate the impact of the input condition in combination with the model architecture on lower resolution (64 $\rightarrow$ 128) for performance reasons. We condition the model either on a depth map or an RGB-D image. The intuition is that the RGB channels provide further visual glues that are not present in the depth modality and hence improve performance. We further vary number of channels and the diffusion steps $T$. We use two different base architectures; first, we use the UNet3+ from run10 from our depth diffusion experiment (see TABLE \ref{tab:depth_diffusion_ablation}). We refer to this architecture as SR1. We observed slow training performance and low performance on our metrics. Following the intuition that for a super-resolution task the higher levels of the U-Net are more important (where the higher spatial information is present and less semantic information has been extracted), we shift the number of channels from deeper to higher levels and use the U-Net instead of the UNet3+ to remove the connections from various levels. We refer to the adapted architecture as SR2.

\begin{table}[t]
\tiny
\centering
\caption{Ablation study for our depth super-resolution model where we vary the model base architecture, the input condition format, the base dimension of the model and the number diffusion steps T. Models with T=1000 are trained for 350 epochs, models with T=600 are trained for 250 epochs. All models perform super-resolution from 64x64 to 128x128. The detailed architecture of the models can be found in TABLE \ref{tab:experiment_model_description_superres_condition} in the appendix.}
\label{tab:superres_condition_and_hyperparams}
\begin{tabular}{ l | l l c c | r | r | r | r }
\hline
\textbf{Run} & \textbf{Arch.} & \textbf{Cond.} & \textbf{Dim.} & \textbf{T} & \textbf{MAE} $\downarrow$ & \textbf{MSE} $\downarrow$ & \textbf{IoU} $\uparrow$ & \textbf{VLB} $\downarrow$ \\ 
\hline
\hline
sr1 & SR1 & RGB-D & 96 & 1000 & 30.20 & 29.68 & 0.815 & 33.87 \\ 
sr2 & SR1 & Depth & 64 & 1000 & 40.14 & 39.19 & 0.789 & 24.92 \\ 
sr3 & SR1 & RGB-D & 64 & 1000 & 38.70 & 37.28 & 0.753 & 19.48 \\ 
sr4 & SR1 & Depth & 64 & 600 & 30.51 & 28.39 & 0.816 & 20.97 \\ 
sr5 & SR1 & RGB-D & 64 & 600 & 47.32 & 41.25 & 0.772 & 34.93 \\
\hline
sr6 & SR2 & RGB-D & 128 & 1000 & \textbf{3.08} & 2.66 & 0.980 & 10.50 \\ 
sr7 & SR2 & Depth & 128 & 1000 & 5.34 & 5.07 & 0.962 & 10.51 \\ 
sr8 & SR2 & RGB-D & 128 & 600 & 3.26 & 2.78 & 0.979 & 10.34 \\ 
sr9 & SR2 & Depth & 128 & 600 & 5.53 & 5.20 & 0.961 & 10.37 \\ 
sr10 & SR2 & RGB-D & 192 & 1000 & 3.11 & \textbf{2.55} & \textbf{0.981} & \textbf{10.21} \\ 
sr11 & SR2 & Depth & 192 & 1000 & 5.56 & 5.22 & 0.961 & 10.26 \\ 
\hline
\end{tabular}
\end{table}

In our second experiment we ablate the usage of condition input augmentation, i.e. blurring the RGB using a Gaussian filter (see equation \eqref{eq:rgb_aug}) and applying depth noise (see equation \eqref{eq:depth_aug}). We report the performance on the metrics in TABLE \ref{tab:superres_augmentation}. We used the model from run sr6, which is similar to the best run sr10, but requires less compute and hence we can iterate faster. We find that the metrics on our test dataset are slightly worse using any of the augmentations and is worst when applying both. However, we observe that the visual quality of generated samples are better when we input the diffused depth output from our depth diffusion network instead of the GT from the test set. We hypothesize that this might be due to the noisier depth of diffused samples and for those samples, the depth noise augmentation is beneficial.

\begin{table}[t]
\tiny
\centering
\caption{Ablation study of the augmentation method for our depth super-resolution model. We use model sr6 from TABLE \ref{tab:superres_condition_and_hyperparams} as our baseline, since it is similar to sr10 but requires less compute, so we can iterate faster. The detailed architecture of the models can be found in TABLE \ref{tab:experiment_model_description_superres_augmentation} in the appendix.}
\label{tab:superres_augmentation}
\begin{tabular}{ l | c c | r | r | r | r  }
\hline
\textbf{Run} & \textbf{RGB blur} & \textbf{Depth Noise} & \textbf{MAE} $\downarrow$ & \textbf{MSE} $\downarrow$ & \textbf{IoU} $\uparrow$ & \textbf{VLB} $\downarrow$ \\ 
\hline
\hline
sr6 & \textbf{x} & \textbf{x} & \textbf{3.08} & \textbf{2.66} & \textbf{0.980} & 10.50 \\ 
\hline
sr61 & \checkmark & \textbf{x} & 3.91 & 2.68 & \textbf{0.980} & 10.47 \\ 
sr62 & \textbf{x} & \checkmark & 4.03 & 2.70 & \textbf{0.980} & \textbf{10.44} \\ 
sr63 & \checkmark & \checkmark & 5.60 & 5.34 & 0.960 & 10.49 \\ 
\hline
\end{tabular}
\end{table}

Finally, we transfer our findings onto training models for higher resolutions, i.e. 64x64 $\rightarrow$ 256x256 and report the performance on the metrics in TABLE \ref{tab:superres_high_res}. We started with run sr12, for which we used the same channel multipliers and number of layers as for sr6. Since attention is a costly operation, we only used it at the same resolutions (i.e. 32x32 and 16x16, because model sr12 does not have 8x8 due to higher input resolution). In run sr121 we added a third level of attention, which kept the VLB roughly equal but improved the IoU, MAE and MSE. With run sr122 we added an extra stage to our U-Net architecture, to verify our hypothesis that lower resolutions do not contribute significantly to the performance of the super-resolution model. The metrics are negligibly better than for sr121 which supports our hypothesis. Based on our results from run sr10, we wanted to test with a base dimensionality of 192 for our higher resolution super-resolution model. We only see minor improvement comparing run sr13 to run sr121. With even worse results for sr131. Since the number of parameters more than doubled from 71M to 161M we think that our dataset is too small and the model overfits on the train set. We consider run sr121 best and select it as final model. 

\begin{table}[t]
\tiny
\centering
\caption{Ablation study for our depth super-resolution model of the target resolution 64x64 $\rightarrow$ 256x256. We vary the models' base dimension, the number of stages and its respective base dim multiplier and the respective resolutions where we perform attention. The detailed architecture of the models can be found in TABLE \ref{tab:experiment_model_description_superres_augmentation} in the appendix.}
\label{tab:superres_high_res}
\begin{tabular}{ l | l l l | r | r | r | r }
\hline
\textbf{Run} & \textbf{Dim.} & \textbf{Mult.} & \textbf{Att.Res.} & \textbf{MAE} $\downarrow$ & \textbf{MSE} $\downarrow$ & \textbf{IoU} $\uparrow$ & \textbf{VLB} $\downarrow$ \\ 
\hline
\hline
sr12  & 128 & 1/1/2/2/4   & 16/32    & 7.73 & 7.22 & 0.945 & 10.34 \\ 
sr121 & 128 & 1/1/2/2/4   & 16/32/64 & 4.33 & 3.02 & 0.977 & 10.37 \\ 
sr122 & 128 & 1/1/2/2/4/4 & 8/16/32  & 4.30 & 3.00 & 0.977 & 10.37 \\ 
sr13  & 192 & 1/1/2/2/4   & 16/32    & \textbf{3.54} & \textbf{2.89} & \textbf{0.978} & 10.06 \\ 
sr131 & 192 & 1/1/2/2/4   & 16/32/64 & 17.98 & 6.04 & 0.821 & \textbf{9.83} \\
\hline
\end{tabular}
\end{table}

\begin{figure}[t]
  \centering
  \includegraphics[width=0.99\linewidth]{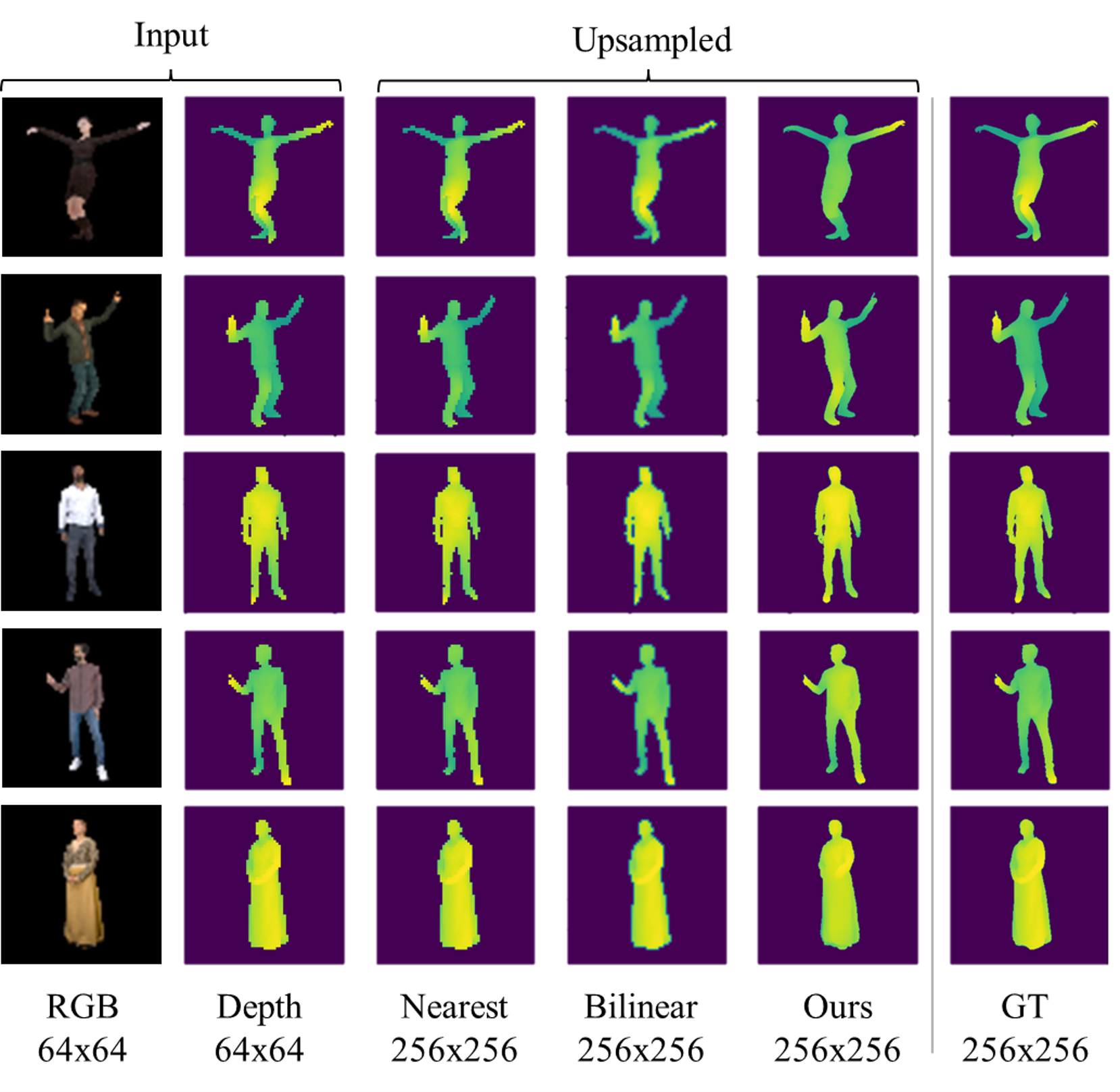}
  \caption{Comparison of our diffusion based depth super-resolution model sr121 against other resampling techniques and the ground truth.}
  \label{fig:superres_method_comparison}
\end{figure}

\begin{figure}[t]
  \centering
  \includegraphics[width=0.99\linewidth]{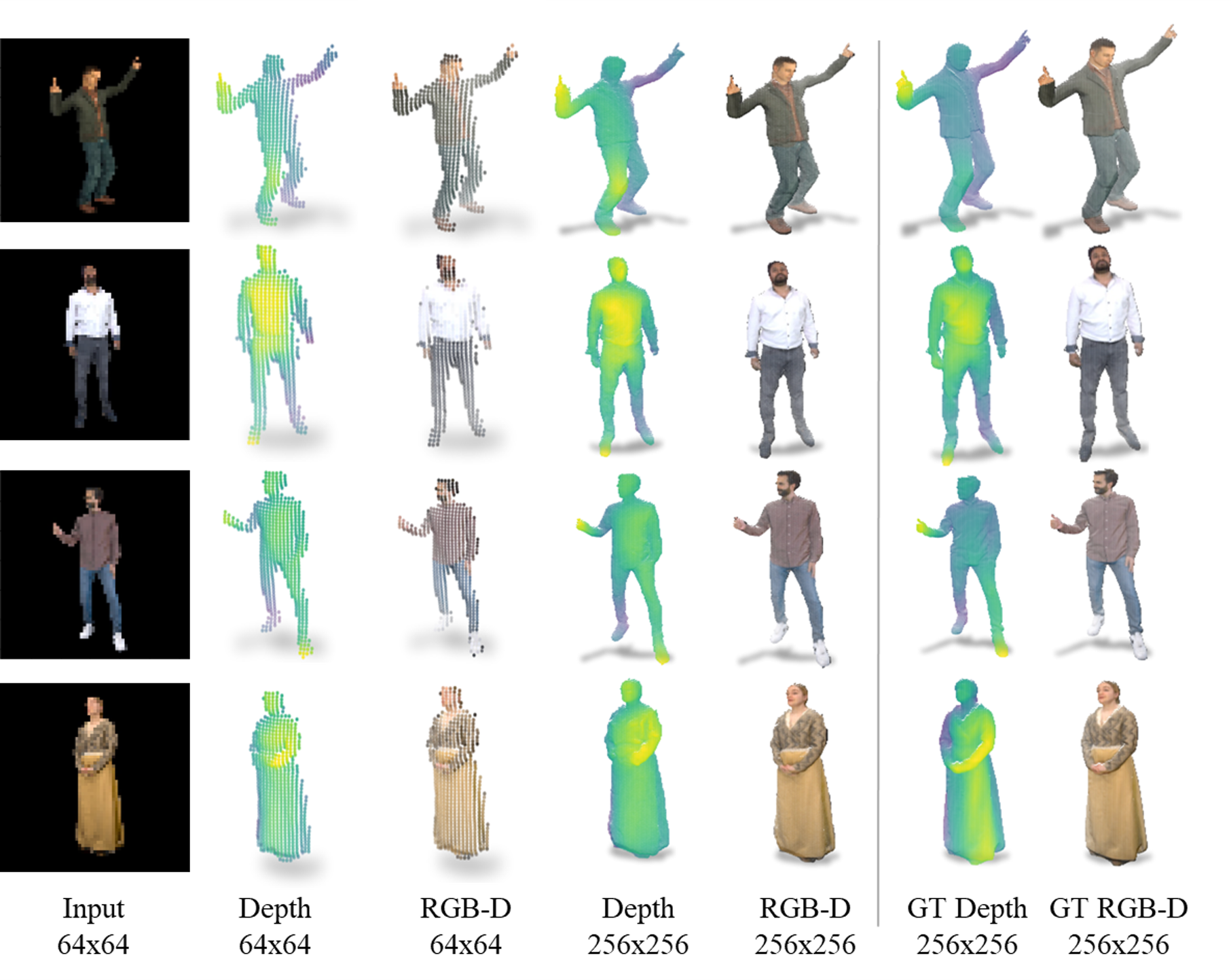}
  \caption{Input and output of our \modelname{} framework at various stages compared to the ground truth. We use model dd10 for the depth diffusion and sr121 for the depth super-resolution}
  \label{fig:rgb_d_fusion_pipeline_output}
\end{figure}

Fig. \ref{fig:superres_method_comparison} depicts a visual comparison of our super-resolution model sr121 compared against nearest neighbor upsampling, bilinear upsampling and the ground truth. In Fig. \ref{fig:rgb_d_fusion_pipeline_output} we show the generated outputs at each stage of our \modelname{} framework.

\begin{figure*}[t]
  \centering
  \includegraphics[width=0.99\linewidth]{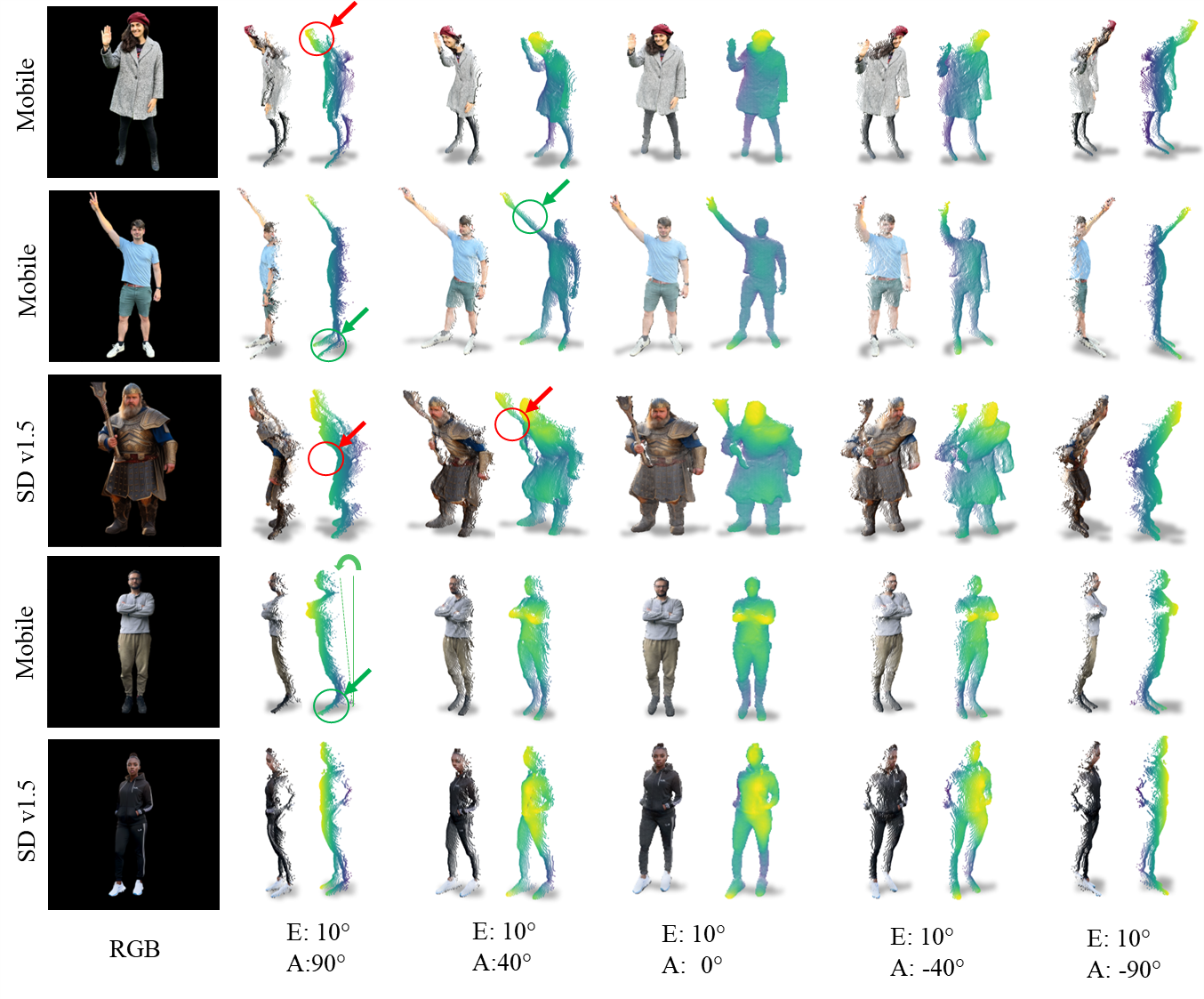}
  \caption{Perspective depth and RGB-D output of our \modelname{} framework conditioned on RGB images obtained from either mobile cameras or images generated with stable diffusion v1.5. Results are illustrated from different viewpoints, i.e. different elevation angles E and azimuth angles A. Errors and distortions are highlighted in red, while distortions related to perspective projection are highlighted in green.}
  \label{fig:in_the_wild}
\end{figure*}

\begin{figure*}[t]
  \centering
  \caption{Outputs of the \modelname{} framework for the given input images. Original images are taken with a variety of single-view mobile cameras.}
  \includegraphics[width=0.95\textwidth]{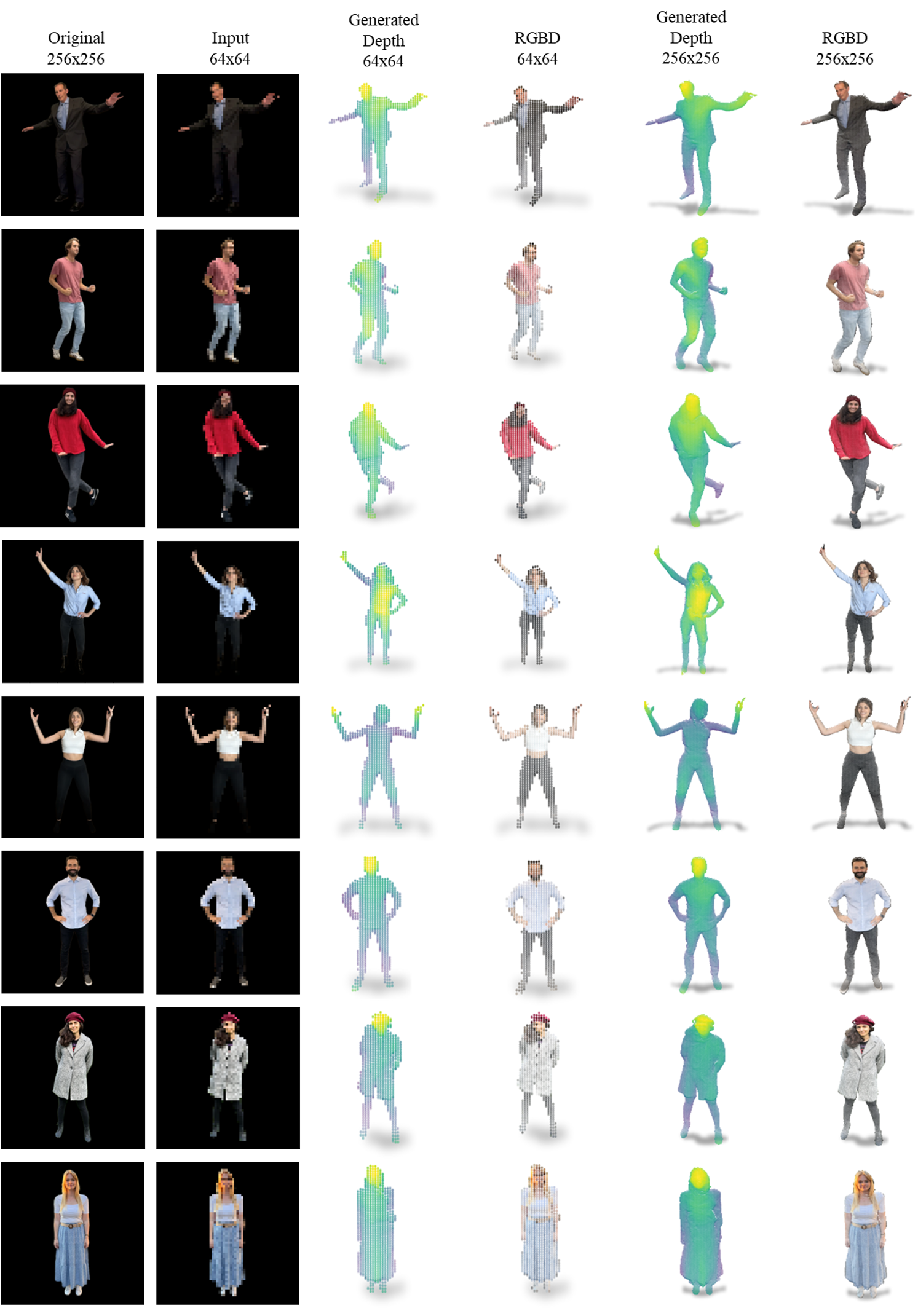}
  \label{fig:rgb_d_fusion_wild_rgbd_2}
\end{figure*}

\begin{figure*}[t]
  \centering
  \caption{Outputs of the \modelname{} framework for the given input images. Original images are generated using a stable diffusion model version 1.5 by \cite{rombach_high-resolution_2022}.}
  \includegraphics[width=0.95\textwidth]{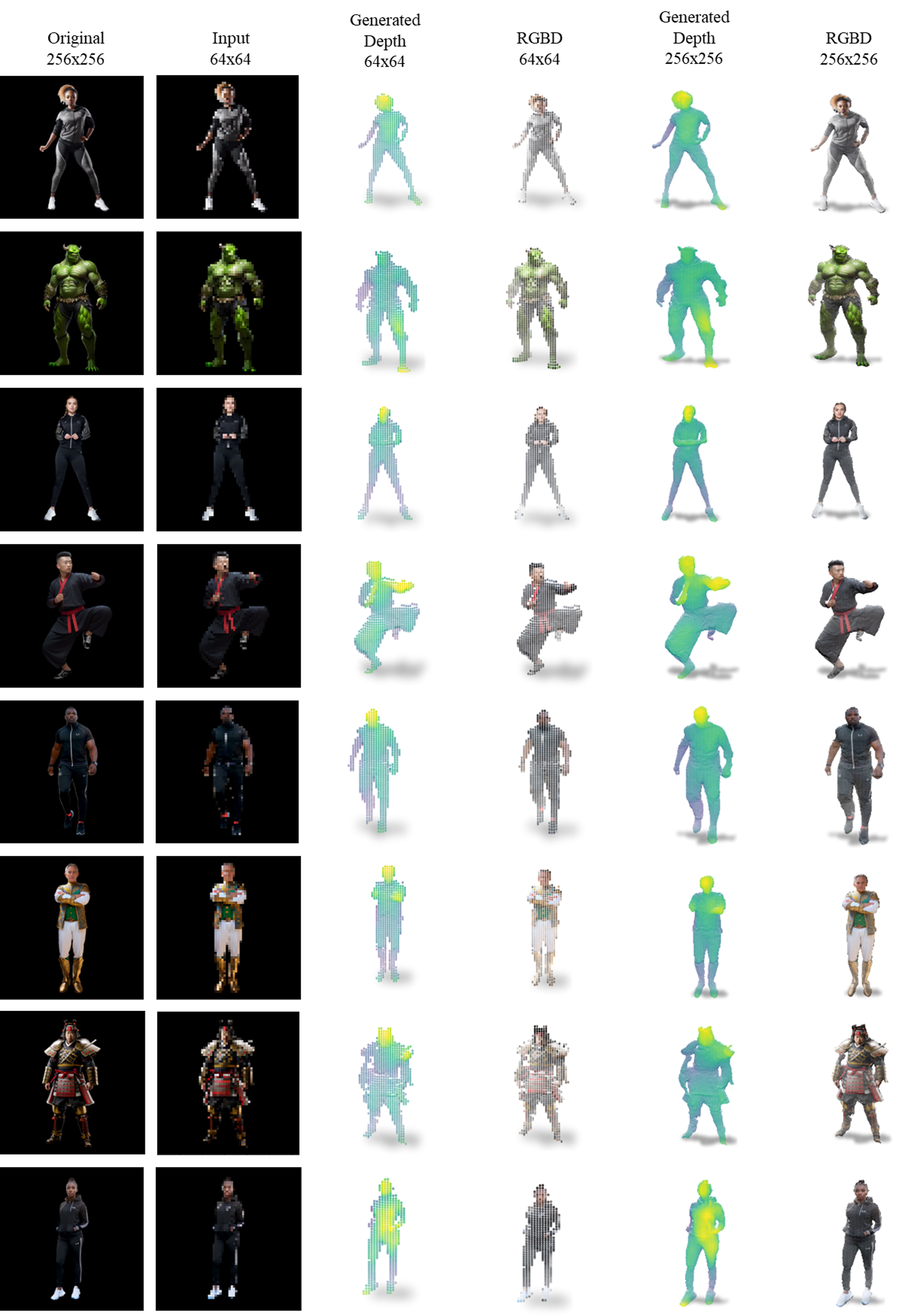}
  \label{fig:rgb_d_fusion_wild_rgbd_3}
\end{figure*}

We further test our \modelname{} framework on "in the wild predictions" where we condition the depth diffusion process on RGB images obtained from different mobile cameras and RGB images generated using stable diffusion v1.5 \citep{rombach_high-resolution_2022}. We aim to investigate how well the framework generalizes to RGB images from a different data domain. Some results are shown in Fig. \ref{fig:in_the_wild} and further results are provided in the appendix in Fig. \ref{fig:rgb_d_fusion_wild_rgbd_1} and Fig. \ref{fig:rgb_d_fusion_wild_rgbd_2} for images obtained from a mobile camera and Fig. \ref{fig:rgb_d_fusion_wild_rgbd_3} for images generated using stable diffusion v1.5.

We observe that for most cases a plausible perspective depth is generated, showing typical distortions such as the subject being tilted towards the camera and lengthen extremities like feet and arms. In some cases the predictions show implausible depths. The super-resolution model on the other hand successfully increases the resolution even of subjects with such implausible depth maps indicating it is robust against domain shifts and solely focusing on increasing the depth maps resolution, which is desirable.
We hypothesize that the implausible depth is caused by a large domain gap between the input data and the training data in at least two aspects. First, a domain gap related to the subject's characteristics like pose, cloths, carried objects as well as lighting and colors in the image. Second, the position and intrinsic parameters of the camera that captured the image are significantly different as those used for rendering our dataset.
\section{Conclusion} \label{sec:conclusion}
In this work, we proposed a comprehensive two-staged framework, namely \modelname{}, to perform monocular depth estimation and depth super-resolution using conditional denoising diffusion probabilistic models. Our depth diffusion model effectively generates depth maps conditioned on low-resolution RGB images in the first stage, while the super-resolution model produces high-resolution depth maps based on a low resolution RGB-D input in the second stage. We introduced a novel depth noise augmentation technique to enhance the robustness of the super-resolution model. Through a series of experiments and ablations we justified design decisions and the effectiveness of our framework and complement those with a variety of generated high resolution RGB-D images.

\subsection{Limitations} \label{subsec:limitations}
While our approach generates high-resolution depth maps, the two-stage approach using two consecutive DDPMs leads to a high demand of resources for sampling and training, especially with increasing image resolutions. We observe that for some "in the wild predictions" the domain gap between input data and training data is to large to generate plausible depth maps indicating a larger dataset might be beneficial. Further, our model outputs a perspective depth which requires a known projection matrix to obtain the depth in any desired coordinate system.

\subsection{Future Work} \label{subsec:future_work}
 Future research might focus on applying faster sampling algorithms and different model architectures like latent diffusion models which might be promising especially for generating depth maps in higher resolutions. Furthermore, one could extent our approach to directly predict the depth in camera or world coordinates or incorporate camera parameters either into the model architecture or in the pre-processing/augmentation stage of the data pipeline. Further, it would be interesting to investigate different representations for the depth data like point clouds or meshes. Finally, one could also imagine, extending our approach to generate occluded regions of the subject, like it's back, combining it to a complete 3D model.

\newpage
\bibliographystyle{plainnat}
\bibliography{main}

\clearpage
\section*{Author biographies}

\setlength\intextsep{0pt} 
\begin{wrapfigure}{l}{25mm} 
    \includegraphics[width=1in,height=1.25in,clip,keepaspectratio]{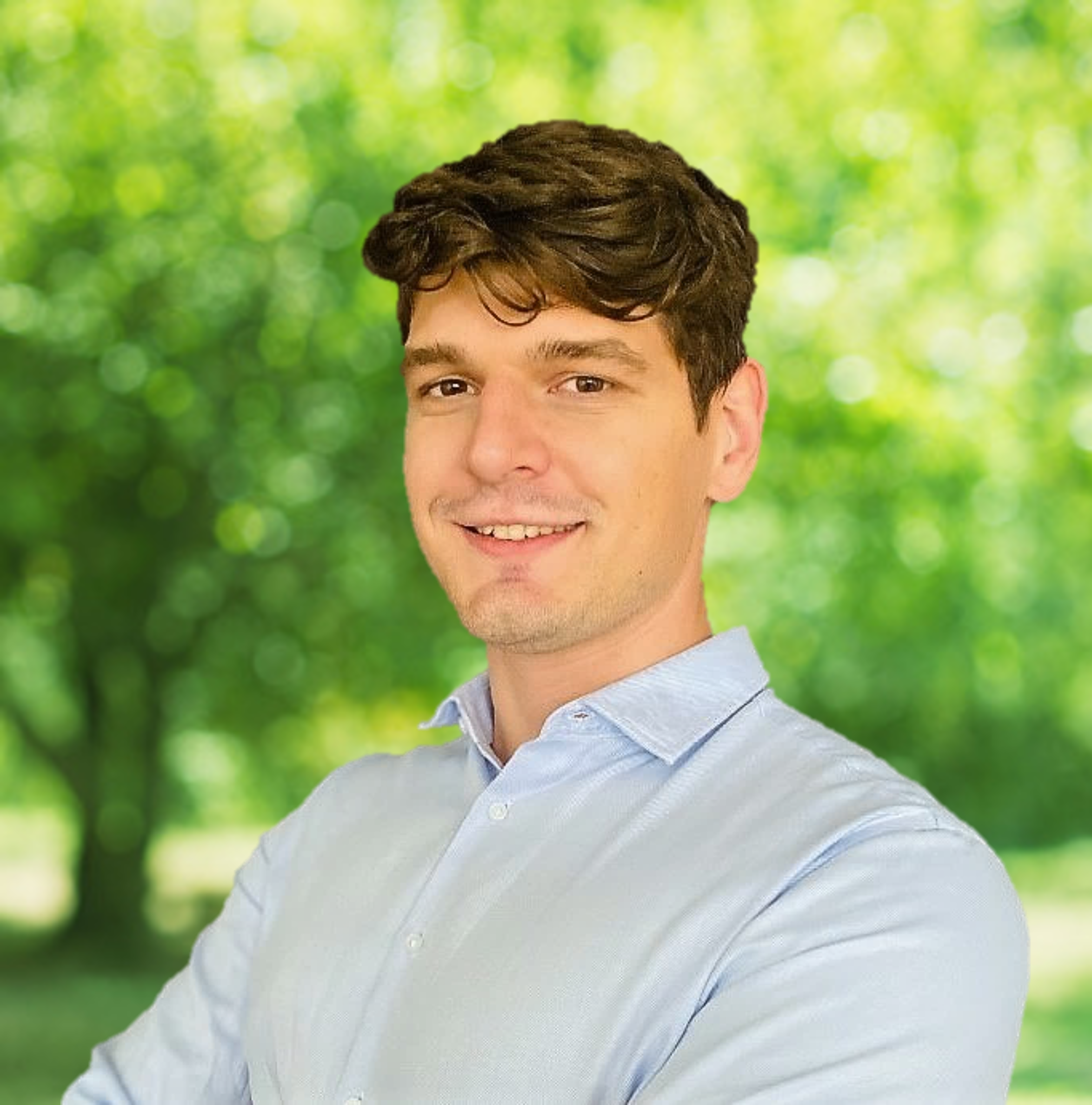}
\end{wrapfigure}\par
\noindent \textbf{Sascha Kirch} is a doctoral student at UNED, Spain. His research focuses on self-supervised multi-modal generative deeplearning. He received his M.Sc. degree in Electronic Systems for Communication and Information from UNED, Spain. He received his B.Eng. degree in electrical engineering from the Cooperative State University Baden-Wuerttemberg (DHBW), Germany. Sascha is member of IEEE’s honor society Eta Kappa Nu as part of the chapter Nu Alpha.\\

\setlength\intextsep{0pt} 
\begin{wrapfigure}{l}{25mm} 
    \includegraphics[width=1in,height=1.25in,clip,keepaspectratio]{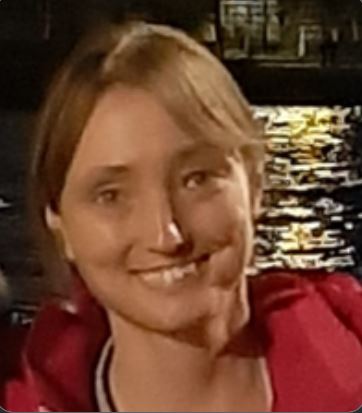}
\end{wrapfigure}\par
\noindent \textbf{Valeria Olyunina} is a 3D Computer Vision Engineer at Volograms Ltd. Her work is centered around volumetric reconstruction of people from video, including research into AI-generated shape estimation and other AI applicable to the subject. She received her M.Sc. degree in Computer Science specialising in Augmented Reality from Trinity College Dublin, Ireland. She also has Postgraduate Diploma in Mathematical modelling and Numerical Solutions from University College Cork, Ireland.\\

\setlength\intextsep{0pt} 
\begin{wrapfigure}{l}{25mm} 
    \includegraphics[width=1in,height=1.25in,clip,keepaspectratio]{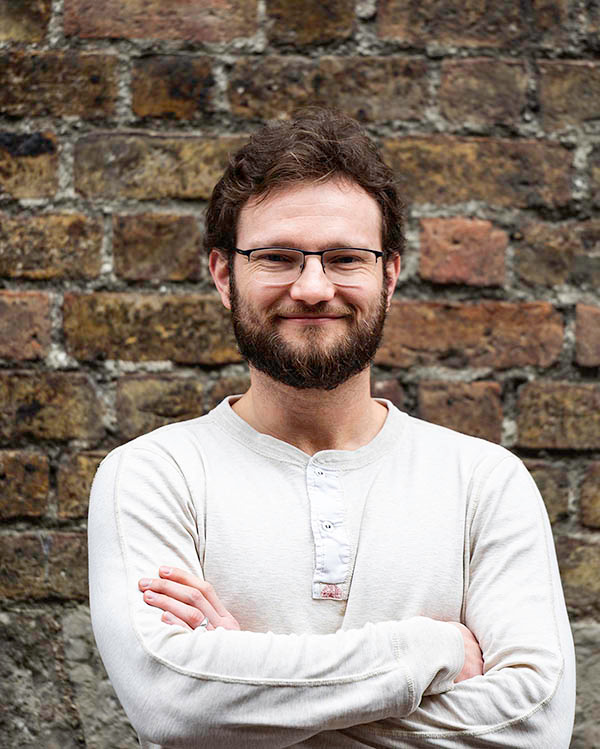}
\end{wrapfigure}\par
\noindent \textbf{Jan Ondřej} is Co-Founder and CTO of Volograms, where he has been since 2018. Previously, he was a postdoctoral researcher at Trinity College Dublin and Disney Research Los Angeles. He obtained M.Sc. in Computer Science in 2007 from Czech Technical University in Prague and his Ph.D. in 2011 from INRIA Rennes, in France. Since 2008 he worked as a researcher in several national and European projects related to volumetric video, animation of virtual humans and crowds, and application of VR/AR technologies.\\

\setlength\intextsep{0pt} 
\begin{wrapfigure}{l}{25mm} 
    \includegraphics[width=1in,height=1.25in,clip,keepaspectratio]{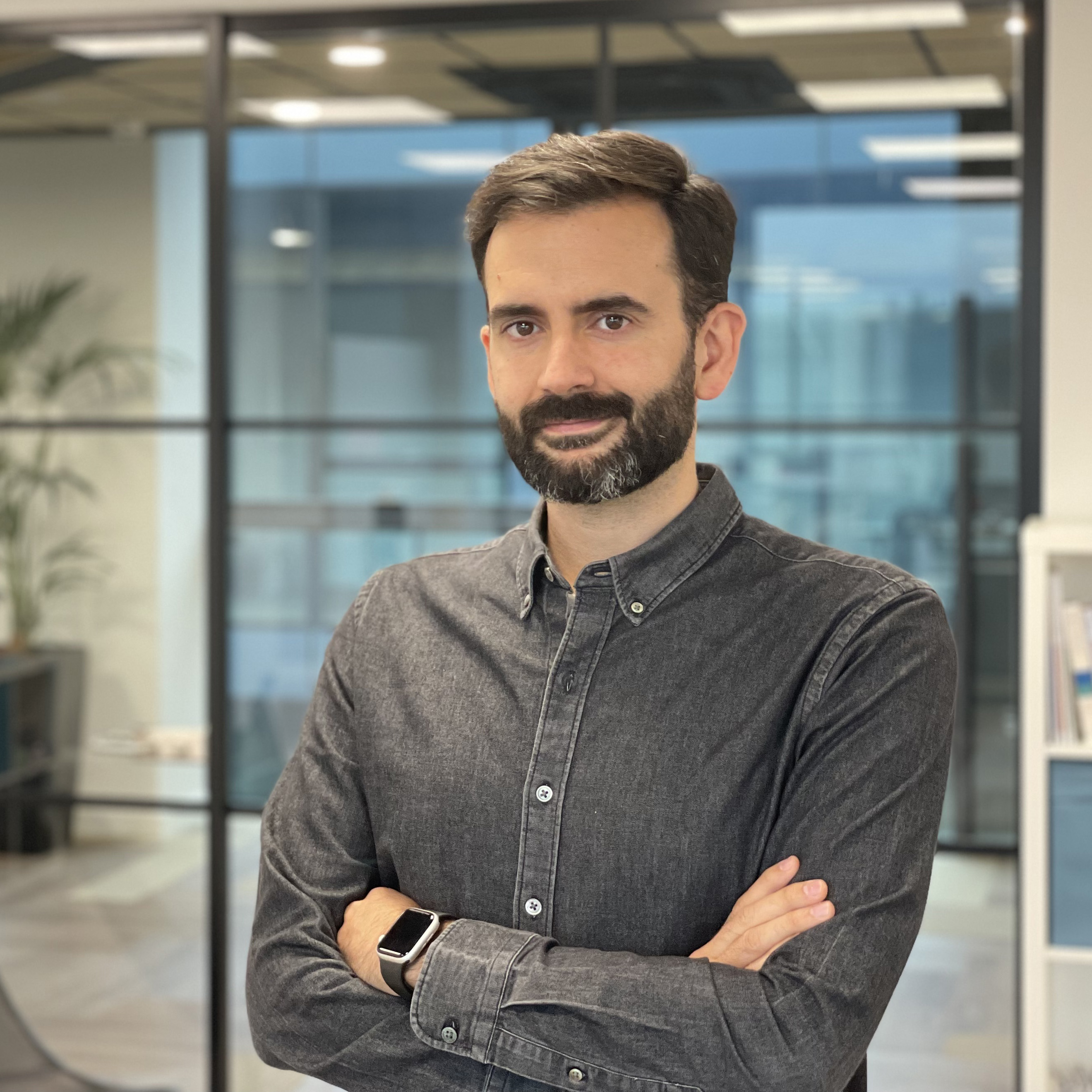}
\end{wrapfigure}\par
\noindent \textbf{Rafael Pagés} is Co-Founder and CEO of Volograms, a startup bringing 3D reconstruction technologies to everyone. He received the Telecommunications Engineering degree (Integrated B.Sc.-M.Sc. accredited by ABET) in 2010, and PhD in Communication Technologies and Systems degree in 2016, both from Technical University of Madrid (UPM), in Spain. Rafael was member of the Image Processing Group at UPM and did his post-doctoral research at Trinity College Dublin. His research interests include 3D reconstruction, volumetric video, and computer vision.\\

\newpage 
\setlength\intextsep{0pt} 
\begin{wrapfigure}{l}{25mm} 
    \includegraphics[width=1in,height=1.25in,clip,keepaspectratio]{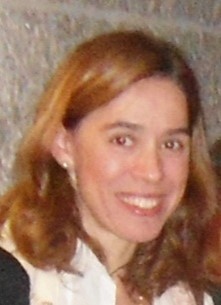}
\end{wrapfigure}\par
\noindent \textbf{Clara Pérez-Molina} received her M.Sc. degree in Physics from the Complutense University in Madrid and her PhD in Industrial Engineering from the Spanish University for Distance Education (UNED). She has worked as researcher in several National and European Projects and has published different technical reports and research articles for International Journals and Conferences, as well as several teaching books. She is currently an Associate Professor with tenure of the Electrical and Computer Engineering Department at UNED.
Her research activities are centered on Educational Competences and Technology Enhanced Learning applied to Higher Education in addition to Renewable Energy Management and Artificial Intelligence techniques. She is senior member of the IEEE.\\

\setlength\intextsep{0pt} 
\begin{wrapfigure}{l}{25mm} 
    \includegraphics[width=1in,height=1.25in,clip,keepaspectratio]{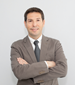}
\end{wrapfigure}\par
\noindent \textbf{Sergio Martín} is Associate Professor at UNED (National University for Distance Education, Spain). He is PhD by the Electrical and Computer Engineering Department of the Industrial Engineering School of UNED. He is Computer Engineer in Distributed Applications and Systems by the Carlos III University of Madrid. He teaches subjects related to microelectronics and digital electronics since 2007 in the Industrial Engineering School of UNED. He has participated since 2002 in national and international research projects related to mobile devices, ambient intelligence, and location-based technologies as well as in projects related to "e-learning", virtual and remote labs, and new technologies applied to distance education. He has published more than 200 papers both in international journals and conferences. He is IEEE senior member.\\

\clearpage
\label{sec:appendix}

\begin{table*}[t]
\tiny
\centering
\caption{Detailed hyperparameters for depth diffusion models used for experiments in TABLE \ref{tab:depth_diffusion_ablation}}
\label{tab:experiment_model_description_depth_diffusion}
\begin{tabular}{ l || r r r r r r r r r r }
\textbf{Run} & \textbf{dd1} & \textbf{dd2} & \textbf{dd3} & \textbf{dd4} & \textbf{dd5} & \textbf{dd6} & \textbf{dd7} & \textbf{dd8} & \textbf{dd9} & \textbf{dd10} \\ 
\hline
\hline
 Model & UNet & UNet3+ & UNet3+ & UNet3+ & UNet3+ & UNet & UNet & UNet3+ & UNet3+ & UNet3+\\
 Diffusion Steps & 600 & 600 & 600 & 600 & 600 & 600 & 600 & 600 & 600 & 600\\
 Condition Input & RGB & RGB & RGB & RGB & RGB & RGB & RGB & RGB & RGB & RGB \\
 Noise Schedule & cosine & cosine & cosine & cosine & cosine & cosine & cosine & cosine & cosine & cosine \\
 Model Size & 41M & 42M & 42M & 42M & 42M & 41M & 41M & 42M & 55M & 55M \\
 Base Dim & 64 & 64 & 64 & 64 & 64 & 64 & 64 & 64 & 64 & 64 \\
 Base Dim Mult. & 1/2/4/8 & 1/2/4/8 & 1/2/4/8 & 1/2/4/8 & 1/2/4/8 & 1/2/4/8 & 1/2/4/8 & 1/2/4/8 & 1/2/4/8 & 1/2/4/8 \\
 \# Blocks & 1/1/1/1 & 1/1/1/1 & 1/1/1/1 & 1/1/1/1 & 1/1/1/1 & 1/1/1/1 & 1/1/1/1 & 1/1/1/1 & 1/1/1/1 & 1/1/1/1 \\
 \# ResNet Blocks & 2/2/2/2 & 2/2/2/2 & 2/2/2/2 & 2/2/2/2 & 2/2/2/2 & 2/2/2/2 & 2/2/2/2 & 2/2/2/2 & 2/2/12/2 & 2/2/12/2 \\
 Stoch. Depth & 0/0/0/0 & 0/0/0/0 & 0/0/0/0 & 0/0/0/0 & 0/0/0/0 & 0/0/0/0 & 0/0/0/0 & 0/0/0/0 & 0/0/0/0 & 0.1/0.1/0.5/0.1 \\
 Input Resolution & 64x64 & 64x64 & 64x64 & 64x64 & 64x64 & 64x64 & 64x64 & 64x64 & 64x64 & 64x64 \\
 Output Resolution & 64x64 & 64x64 & 128x128 & 64x64 & 64x64 & 64x64 & 64x64 & 64x64 & 64x64 & 64x64 \\
 Batch size & 128 & 128 & 128 & 128 & 64 & 24 & 24 & 12 & 12 & 12 \\
 Variance & fix & fix & learned & learned & learned & learned & learned & learned & learned & learned \\
 Loss Weighting & simple & simple & simple & P2 & P2 & P2 & simple & simple & simple & simple \\
 $L_r$ & 4e-5 & 4e-5 & 4e-5 & 4e-5 & 6e-5 & 4e-5 & 4e-5 & 4e-5 & 4e-5 & 4e-5 \\
 $L_r$ Schedule & \textbf{x} & \textbf{x} & \textbf{x} & \textbf{x} & \checkmark & \checkmark & \checkmark & \checkmark & \checkmark & \checkmark \\
 Att. Resolution & 64/32/16/8 & 64/32/16/8 & 64/32/16/8 & 64/32/16/8 & 64/32/16/8 & 64/32/16/8 & 64/32/16/8 & 64/32/16/8 & 64/32/16/8 & 64/32/16/8 \\
 Att. Heads & 8 & 8 & 8 & 8 & 8 & 8 & 8 & 8 & 8 & 8\\
 Att. Heads channels & 32 & 32 & 32 & 32 & 32 & 32 & 32 & 32 & 32 & 32\\
 GN Group Size & 8 & 8 & 8 & 8 & 8 & 8 & 8 & 8 & 8 & 8 \\
 Dropout & 0.1 & 0.1 & 0.1 & 0.1 & 0.1 & 0.1 & 0.1 & 0.1 & 0.1 & 0.1 \\
\hline
\end{tabular}
\end{table*}

\begin{table*}[t]
\tiny
\centering
\caption{Super-resolution models used for experiments in TABLE \ref{tab:superres_condition_and_hyperparams}}
\label{tab:experiment_model_description_superres_condition}
\begin{tabular}{ l || r r r r r | r r r r r r }
\textbf{Run} & \textbf{sr1} & \textbf{sr2} & \textbf{sr3} & \textbf{sr4} & \textbf{sr5} & \textbf{sr6} & \textbf{sr7} & \textbf{sr8} & \textbf{sr9} & \textbf{sr10} & \textbf{sr11} \\ 
\hline
\hline
 Model & UNet3+ & UNet3+ & UNet3+ & UNet3+ & UNet3+ & UNet & UNet & UNet & UNet & UNet & UNet\\
 Architecture & SR1 & SR1 & SR1 & SR1 & SR1 & SR2 & SR2 & SR2 & SR2 & SR2 & SR2\\
 Diffusion Steps & 1000 & 1000 & 1000 & 600 & 600 & 1000 & 1000 & 600 & 600 & 1000 & 1000 \\
 Condition Input & RGB-D & Depth & RGB-D & Depth & RGB-D & RGB-D & Depth & RGB-D & Depth & RGB-D & Depth \\
 Noise Schedule & cosine & cosine & cosine & cosine & cosine & cosine & cosine & cosine & cosine & cosine & cosine \\
 Model Size & 153M & 69M & 69M & 69M & 69M & 72M & 72M & 72M & 72M & 161M & 161M \\
 Base Dim & 96 & 64 & 64 & 64 & 64 & 128 & 128 & 128 & 128 & 192 & 192 \\
 Base Dim Mult. & 1/2/4/4/8 & 1/2/4/4/8 & 1/2/4/4/8 & 1/2/4/4/8 & 1/2/4/4/8 & 1/1/2/2/4 & 1/1/2/2/4 & 1/1/2/2/4 & 1/1/2/2/4 & 1/1/2/2/4 & 1/1/2/2/4 \\
 \# Blocks & 1/1/1/1/1 & 1/1/1/1/1 & 1/1/1/1/1 & 1/1/1/1/1 & 1/1/1/1/1 & 1/1/1/1/1 & 1/1/1/1/1 & 1/1/1/1/1 & 1/1/1/1/1 & 1/1/1/1/1 & 1/1/1/1/1 \\
 \# ResNet Blocks & 2/2/2/12/2 & 2/2/2/12/2 & 2/2/2/12/2 & 2/2/2/12/2 & 2/2/2/12/2 & 3/3/3/3/3 & 3/3/3/3/3 & 3/3/3/3/3 & 3/3/3/3/3 & 3/3/3/3/3 & 3/3/3/3/3 \\
 Stoch. Depth & 0.1/0.1/0.1/0.5/0.1 & 0.1/0.1/0.1/0.5/0.1 & 0.1/0.1/0.1/0.5/0.1 & 0.1/0.1/0.1/0.5/0.1 & 0.1/0.1/0.1/0.5/0.1 & 0/0/0/0/0 & 0/0/0/0/0 & 0/0/0/0/0 & 0/0/0/0/0 & 0/0/0/0/0 & 0/0/0/0/0 \\
 Input Resolution & 64x64 & 64x64 & 64x64 & 64x64 & 64x64 & 64x64 & 64x64 & 64x64 & 64x64 & 64x64 & 64x64 \\
 Output Resolution & 128x128 & 128x128 & 128x128 & 128x128 & 128x128 & 128x128 & 128x128 & 128x128 & 128x128 & 128x128 & 128x128 \\
 Batch size & 8 & 16 & 16 & 16 & 16 & 16 & 16 & 16 & 16 & 8 & 8 \\
 Variance & learned & learned & learned & learned & learned & fix & fix & fix & fix & fix & fix \\
 Loss Weighting & P2 & P2 & P2 & P2 & P2 & simple & simple & simple & simple & simple & simple \\
 $L_r$ & 1.5e-5 & 1e-5 & 1e-5 & 1e-5 & 1e-5 & 5e-5 & 5e-5 & 5e-5 & 5e-5 & 1e-4 & 1e-4 \\
 $L_r$ Schedule & \checkmark & \checkmark & \checkmark & \checkmark & \checkmark & \checkmark & \checkmark & \checkmark & \checkmark & \checkmark & \checkmark \\
 Att. Resolution & 32/16/8 & 32/16/8 & 32/16/8 & 32/16/8 & 32/16/8 & 32/16/8 & 32/16/8 & 32/16/8 & 32/16/8 & 32/16/8 & 32/16/8 \\
 Att. Heads & 8 & 8 & 8 & 8 & 8 & 8 & 8 & 8 & 8 & 8 & 8 \\
 Att. Heads channels & 32 & 32 & 32 & 32 & 32 & 32 & 32 & 32 & 32 & 32 & 32 \\
 GN Group Size & 8 & 8 & 8 & 8 & 8 & 8 & 8 & 8 & 8 & 8 & 8 \\
 Dropout & 0.1 & 0.1 & 0.1 & 0.1 & 0.1 & 0.1 & 0.1 & 0.1 & 0.1 & 0.1 & 0.1 \\
\hline
\end{tabular}
\end{table*}

\begin{table*}[t]
\tiny
\centering
\caption{Super-resolution models used for experiments in TABLE \ref{tab:superres_augmentation} and \ref{tab:superres_high_res}}
\label{tab:experiment_model_description_superres_augmentation}
\begin{tabular}{ l || r r r || r r r | r r }
\textbf{Run} & \textbf{sr61} & \textbf{sr62} & \textbf{sr63} & \textbf{sr12} & \textbf{sr121} & \textbf{sr122} & \textbf{sr13} & \textbf{sr131} \\ 
\hline
\hline
 Model & UNet & UNet & UNet & UNet & UNet & UNet & UNet & UNet\\
 Architecture & SR2 & SR2 & SR2 & SR2 & SR2 & SR2 & SR2 & SR2\\
 Diffusion Steps & 1000 & 1000 & 1000 & 1000 & 1000 & 1000 & 1000 & 1000 \\
 Condition Input & RGB-D & RGB-D & RGB-D & RGB-D & RGB-D & RGB-D & RGB-D & RGB-D \\
 Noise Schedule & cosine & cosine & cosine & cosine & cosine & cosine & cosine & cosine \\
 Model Size & 72M & 72M & 72M & 72M & 72M & 131M & 161M & 161M \\
 Base Dim & 128 & 128 & 128 & 128 & 128 & 128 & 192 & 192 \\
 Base Dim Mult. & 1/1/2/2/4 & 1/1/2/2/4 & 1/1/2/2/4 & 1/1/2/2/4 & 1/1/2/2/4 & 1/1/2/2/4/4 & 1/1/2/2/4 & 1/1/2/2/4 \\
 \# Blocks & 1/1/1/1/1 & 1/1/1/1/1 & 1/1/1/1/1/1 & 1/1/1/1/1 & 1/1/1/1/1 & 1/1/1/1/1/1 & 1/1/1/1/1 & 1/1/1/1/1 \\
 \# ResNet Blocks & 3/3/3/3/3 & 3/3/3/3/3 & 3/3/3/3/3 & 3/3/3/3/3 & 3/3/3/3/3 & 3/3/3/3/3/3 & 3/3/3/3/3 & 3/3/3/3/3 \\
 Stoch. Depth & 0/0/0/0/0 & 0/0/0/0/0  & 0/0/0/0/0 & 0/0/0/0/0 & 0/0/0/0/0 & 0/0/0/0/0/0 & 0/0/0/0/0 & 0/0/0/0/0 \\
 Input Resolution & 64x64 & 64x64 & 64x64 & 64x64 & 64x64 & 64x64 & 64x64 & 64x64 \\
 Output Resolution & 128x128 & 128x128 & 128x128 & 256x256 & 256x256 & 256x256 & 256x256 & 256x256 \\
 Batch size & 16 & 16 & 16 & 4 & 4 & 4 & 2 & 2\\
 Variance & fix & fix & fix & fix & fix & fix & fix & fix \\
 Loss Weighting & simple & simple & simple & simple & simple & simple & simple & simple \\
 $L_r$ & 5e-5 & 5e-5 & 5e-5 & 3e-5 & 3e-5 & 3e-5 & 5e-5 & 5e-5 \\
 $L_r$ Schedule & \checkmark & \checkmark & \checkmark & \checkmark & \checkmark & \checkmark & \checkmark & \checkmark \\
 Att. Resolution & 32/16/8 & 32/16/8 & 32/16/8 & 32/16 & 64/32/16 & 32/16/8 & 32/16 & 64/32/16 \\
 Att. Heads & 8 & 8 & 8 & 8 & 8 & 8 & 8 & 8 \\
 Att. Heads channels & 32 & 32 & 32 & 32 & 32 & 32 & 32 & 32 \\
 GN Group Size & 8 & 8 & 8 & 8 & 8 & 8 & 8 & 8\\
 Dropout & 0.1 & 0.1 & 0.1 & 0.1 & 0.1 & 0.1 & 0.1 & 0.1 \\
 Aug. Depth Noise & \textbf{x} & \checkmark & \checkmark & \checkmark & \checkmark & \checkmark & \checkmark & \checkmark \\
 Aug. RGB Blur & \checkmark & \textbf{x} & \checkmark & \textbf{x} & \textbf{x} & \textbf{x} & \textbf{x} & \textbf{x} \\
\hline
\end{tabular}
\end{table*}


\begin{figure*}[t]
  \centering
  \caption{Outputs of the \modelname{} framework for the given input images. Original images are taken with a variety of single-view mobile cameras.}
  \includegraphics[width=0.95\textwidth]{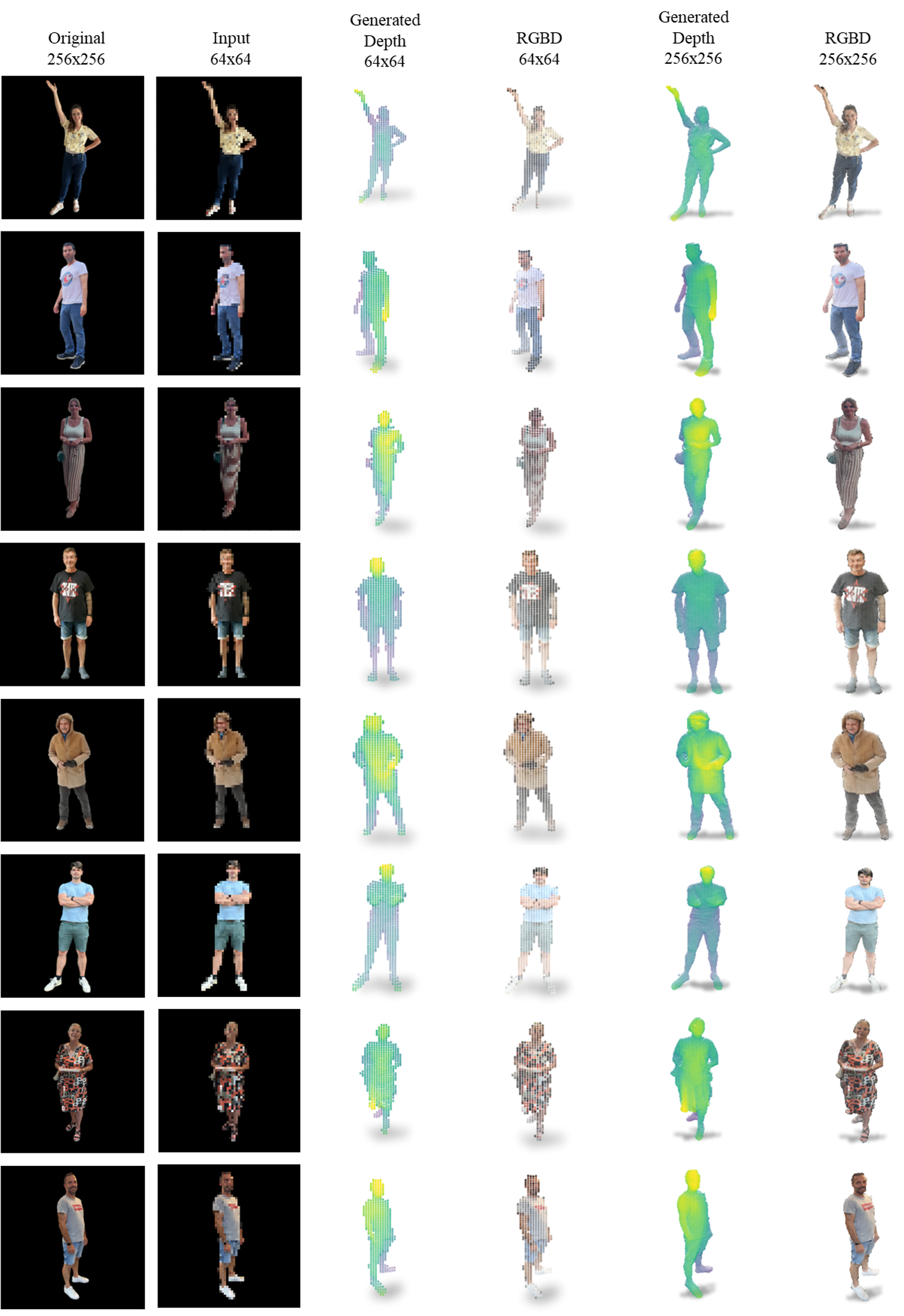}
  \label{fig:rgb_d_fusion_wild_rgbd_1}
\end{figure*}

\end{document}